# An Accelerator for Rule Induction in Fuzzy Rough Theory

Suyun Zhao, Zhigang Dai, Xizhao Wang, *Fellow*, *IEEE*, Peng Ni, Hengheng Luo, Hong Chen, Cuiping Li

*Abstract*—Rule-based classifier, that extract a subset of induced rules to efficiently learn/mine while preserving the discernibility information, plays a crucial role in human-explainable artificial intelligence. However, in this era of big data, rule induction on the whole datasets is computationally intensive. So far, to the best of our knowledge, no known method focusing on accelerating rule induction has been reported. This is first study to consider the acceleration technique to reduce the scale of computation in rule induction. We propose an accelerator for rule induction based on fuzzy rough theory; the accelerator can avoid redundant computation and accelerate the building of a rule classifier. First, a rule induction method based on consistence degree, called Consistence-based Value Reduction (CVR), is proposed and used as basis to accelerate. Second, we introduce a compacted search space termed Key Set, which only contains the key instances required to update the induced rule, to conduct value reduction. The monotonicity of Key Set ensures the feasibility of our accelerator. Third, a rule-induction accelerator is designed based on Key Set, and it is theoretically guaranteed to display the same results as the unaccelerated version. Specifically, the rank preservation property of Key Set ensures consistency between the rule induction achieved by the accelerator and the unaccelerated method. Finally, extensive experiments demonstrate that the proposed accelerator can perform remarkably faster than the unaccelerated rule-based classifier methods, especially on datasets with numerous instances.

*Index Terms*—Accelerator, consistence degree, rule-based classifier, rule extraction, rule induction

## I. Introduction

Fuzzy systems, as one of the most significant advances in computational intelligence, have performed excellent fuzzy modelling capabilities in many data-science scenarios. To date, many mathematical-tool-based fuzzy systems have been widely used in decision making/multi-objective optimization applications. For example, fuzzy programming has been used in flexible-responsive manufacturing/supply system [18,30], and optimization method based on fuzzy credibility theory has been used in applications with fuzzy demand [28]. In addition, some mathematical models with different optimization algorithms have been designed for solving product portfolio problems or fuzzy demand [5,17, 46].

In addition to mathematical-tool-based fuzzy systems, fuzzy-logic-based systems can be adopted to analyze decision making/optimization processes for both research and applications. Recently, rule induction, as an explainable decision making system, has been employed in classification and/or prediction by inducing some rules. Here, each rule contains a conjunctive feature expression and a class node [34,41,42]. An example of rule containing two feature expressions for class C is IF ($x_2$='low') AND ($x_1$<'high') THEN C. This kind of fuzzy systems based on rules, that can explain how outputs are inferred from inputs, allow us to represent knowledge about patterns of interest in an explanatory and understandable manner that can be used by experts [11,16].

Rule-based-classifier, that selects a subset of induced rules to reach efficient mining, is one well-interpretable decision-making system based on rule induction. However, its construction usually involves computationally intensive learning algorithms that require long runtimes [16]. Consequently, the induction of compact and generalized rules in dealing with large-scale datasets has always been a challenging task. Fuzzy rough set (FRS) [3,4,6], as a powerful tool for reducing database dimensionality [10] and building rule-based classifier [7], suffers from the same limitations in dealing with large-scale datasets. Because FRSs, assuming that instances characterized by the same information are indiscernible (similar) [6], must discern all the heterogeneous pairs in the Universe; they produce taxing computation and work less efficiently on large-scale datasets [6,10]. To overcome this drawback, researchers have proposed some heuristic knowledge reduction algorithms including parallel, incremental, and accelerated methods.

Some parallel methods have been proposed to expedite the computation by parallelizing the conventional attribute reduction process based on MapReduce mechanism [20,21]. However, these parallel methods must still consider all the instances in the Universe. Moreover, they cannot avoid performing redundant computations. Some other researchers have proposed incremental attribute-reduction methods [22,24], which mainly focus on handling the dynamic datasets, such as streaming the incoming attributes, instances and/or attribute values. To date, no parallel or incremental methods of

This work is supported by the National Key Research & Develop Plan (2018YFB1004401), NSFC (No. 61732006, 61732011, 61702522, 61976141, 61772536, 61772537, 62072460, 62076245), and Beijing Natural Science Foundation (4212022); it is also partially supported by Hebei Key Laboratory of Machine Learning and Computational Intelligence, Hebei University.
S.Y. Zhao is with Key Laboratory of Data Engineering and Knowledge Engineering (Renmin University of China) and School of Information, Beijing, China (*corresponding author: e-mail: zhao.suyun@yahoo.com).
X.Z. Wang is with Shenzhen University, Shenzhen, China. (e-mail: xizhaowang@ieee.org).
Z.G. Dai, P. Ni, H.H. Luo, H. Chen, C.P. Li are with Key Laboratory of Data Engineering and Knowledge Engineering (Renmin University of China) and School of Information (e-mail: chong@ruc.edu.cn, cuiping_li@263.net).

expediting the FRS-based classifier building have been reported.

Besides the parallel and incremental techniques, researchers have proposed and developed several versions of the accelerator for rough/fuzzy rough attribute reduction [14, 25, 27]. The acceleration technique is an effective way to reduce the scale of knowledge reduction by avoiding redundant computation successively. As a pioneering work, from the perspective of instances, Qian et al. proposed an accelerator called positive approximation for attribute reduction in complete and incomplete data [14,27]. From the perspective of both instances and attributes, Liang et al. [25] introduced an accelerator that could remove insignificant attributes in the process of attribute reduction. More interestingly, Wei et al. [15] accelerated incremental attribute-reduction by compacting a decision table. During attribute reduction, these accelerators remove the redundant instances, that are discerned by the current selected attributes. However, these existing accelerators focus on accelerating attribute selection. So far, the use of accelerators in rough/fuzzy-rough based classifier building has not been reported. This motivated us to propose a fuzzy rough based accelerator to expedite the building of a precise and explainable classifier.

In the rough set theory, a rule-based classifier extracts some induced rules that constitute the minimal-rule-subset retaining the whole discernible information in a decision system. The process of building a rule-based classifier, as shown in Fig. 1.1, has two stages: rule induction from original instances, and rule extraction from induced rules. Currently there exist some heuristic algorithms of rule extraction [12,13,40], which are designed by a certain search strategy, such as forward adding [40] and backward deleting [12,13], to expeditiously build a compact and complete classifier. However, to the best of our knowledge, no study has reported on the acceleration of rule induction. Rule induction, as a unit operation in classifier building, explores the whole universe to induce a compact attribute-value set as the former of the corresponding rule. Thus, conducting rule induction on every instance is computationally intensive and even more infeasible in the case of large-scale data. In this study, we did not reconstruct the search strategy of rule extraction, but accelerate the rule induction of the existing algorithms by compacting their exploration space.

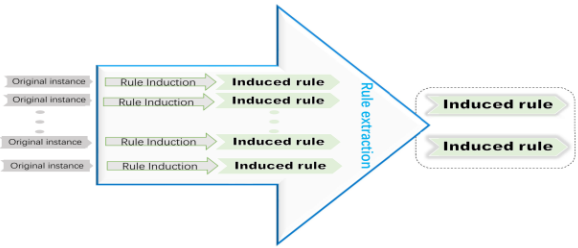

Fig. 1.1: Flow chat of rule-based classifier construction. The process of building rule-based classifier has two stages: rule induction from original instances, and rule extraction from all induced rules.

The consistence degree, fundamental to rule induction, is obtained from the lower approximation by exploring all the Universe and is time-consuming. Many heterogeneous instance pairs, which were already discerned in the process of attribute-value reduction, are still used in the subsequent calculation to find the new informative attribute values. Successively removing those instance pairs, which do not contribute in the calculation of the consistence degree, will accelerate rule reduction. In this study, we first propose a rule induction algorithm based on consistence degree, **CVR**, and exhibit that it is computing intensive. Then we accelerate this rule induction algorithm on a compacted search space, Key Set. This key set only contains the informative instances key to rule induction and then reduce the redundant computation. Third, we propose an accelerator for rule induction based on Key Set. Finally, accelerated rule-based-classifier framework is designed with the accelerated rule induction.

The main contributions of this study are summarized as follows.
- This is the first study to introduce the acceleration techniques into rule induction based on the fuzzy rough theory.
- The search space of rule induction is compacted on Key Set, which only contains the instances key to update the rule reduction. Thus, our rule induction accelerator based on Key Set avoids redundant computation and accelerates the building of a rule classifier.
- The strict mathematical reasoning verifies that our proposed accelerator is reliable. The monotonicity of Key Set ensures the feasibility of our accelerator. The rank preservation property of the key set ensures the rule induction achieved by the accelerator is the same as that achieved by the unaccelerated one.

The remainder of this paper is organized as follows. Section 2 briefly introduce FRSs and a rule-classifier building method, **GFRC**. Section 3 presents an alternative, named **CVR**, to the existing rule-induction algorithm. In Section 4 we present an accelerated version of rule induction, **A-CVR**. Section 5 describes the classifier building framework based on accelerated rule induction. In Section 6, we compare the efficiency and performance of the accelerated rule classifier against to those of unaccelerated and explainable classifiers on the selected UCI/KEEL database. Finally, Section 7 concludes this study.

## II. PRELIMINARIES

In this section, we review some related works on rule induction and cost reduction. Moreover, we briefly review FRS [6, 9, 11] and a state-of-the-art rule induction method proposed in [7].

### A. Related works
#### 1) Methods on Rule Induction

Rule induction is one of the important applications of rough sets. Greco et al. have pointed out that the rules induced by rough sets are more understandable and more applicable for final users because the rules follow a general syntax more closely [33]. A rough set technique has been introduced for solving the problem of mining Pinyin-to-character (PTC) conversion rules [35]. Moreover, there are many works about rule induction based on rough sets' extensions. Wang et al. investigated some fuzzy rules from fuzzy samples based on the rough fuzzy set and fuzzy rough set techniques [7,36]. Huang et al. [39] induce simple dominance-based interval-valued intuitionistic fuzzy rules by using dominance-based rough set model. In addition, Inuiguchi et al. design rule induction algorithms from two decision tables as a basis of rough set analysis of more than one decision table [40]. The four types of decision rules induced in the context of rough sets have provided a generalized description of objects [26].

Furthermore, there are many studies proposed to update the induced rules incrementally. Shan et al proposed a method to update decision rules by updating the decision matrices and decision functions while a new object is added [23]. Huang proposed a rule induction method when objects vary in FRSs [37]. Chen et al. proposed a method of incrementally updating rules based on traditional rough sets when attribute values' coarsening and refining [38]. However, there are no reported investigations on accelerating rule induction on static environment.

*2) Fuzzy Methods on reducing costs*

In past decades, various optimization methods, such as mathematical programming [30,43], genetic algorithms [44], multi-objective search [29,31,32] and Harmony search [28] were employed in problems of arranging instances/allocating resources with the goal of reducing cost. In [28], an optimization method is designed based on credibility theory and a harmony algorithm with random simulation, to solve the proposed mathematical model for routing relief vehicles. In [30], a fuzzy Mixed Integer Linear Programming along with a hybrid genetic algorithm and a Whale optimization algorithm is developed to address the cell formation and inter-cellar scheduling problem in a Cellar Manufacturing System environment. In addition, a hybrid artificial intelligence and robust optimization for the product portfolio problem under return uncertainty is proposed in [32]. These methods usually propose a mathematical model for the real applications and then address them by some optimization algorithms with the goal of cost reduction.

However, the optimization suggested in these studies is time-consuming. Their final goal is to achieve the optimal/suboptimal solutions of resource allocation/instance arrangement, which is distinct from our goal of accelerating rule induction. Accelerator aims to reduce computational costs as well as retaining the learning quality by proper arrangement of involved instances [14]. In accelerating methods, instance arrangement/ resource allocation is just a tool to accelerate rule induction, not the final goal. Thus, the afore-mentioned optimization strategies are not suitable for our accelerating problems. It is worth noting that the goal of the optimization strategies is consistent with that of rule extraction. In another paper, we would like to design a novel method of rule extraction to achieve qualified and minimal rule set, by using certain optimization strategy.

*B. FRSs*

Usually, data are described as one decision table, denoted by $DT = (U, C \cup D)$. Let the Universe, denoted by $U$, be a nonempty set with a finite number of instances $\{x_1, x_2, \ldots, x_n\}$. Each instance $x_i$ in $U$ is described by a nonempty finite set of condition attributes, denoted by $C$, and the set of decision attributes, denoted by $D$. Note that $C \cap D = \emptyset$.

If $A$ is a mapping: $U \to [0,1]$, then $A$ is called a fuzzy set on $U$, $A(x) \in [0,1]$ is the fuzzy membership degree of $x \in U$ belonging to fuzzy set $A$ [8]. If attribute $r \in C$ is such a kind of mapping $U \to [0,1]$, then it is fuzzy. As each continuous attribute can be transferred into a fuzzy one, the decision table with continuous attributes is then called a Fuzzy Decision Table, denoted by $FD$. The FRSs, combining fuzzy sets [8] and rough sets [1,2], can effectively work on $FD$ [9]. Some concepts and properties of FRSs are briefly reviewed in the following. For further details on FRS, refer to [6,9].

In a fuzzy decision table, each attribute subset $B \subseteq C$ corresponds to a similarity relation $\tilde{R}_B(\cdot,\cdot)$, which satisfies, for every $x,y,z \in U$, (1) Reflexivity ($\tilde{R}_B(x,x) = 1$); (2) Symmetry ($\tilde{R}_B(x,y) = \tilde{R}_B(y,x)$); (3) T-transitivity ($\tilde{R}_B(x,y) \geq T(\tilde{R}_B(x,z), \tilde{R}_B(z,y))$), where $T$ is a triangular norm (see the appendix for the definitions of the triangular norm). $\forall x, y \in U$, $\tilde{R}_B(x,y) = \min_{a \in B}(\tilde{R}_a(x,y))$.

FRS was first proposed by Dubois and Prade [3,4], which is defined as follows.

**Definition 2.1.** An **FRS** is an ordered pair $(\underline{R_B}A, \overline{R_B}A)$ of fuzzy set $A$ on $U$ such that for every $x \in U$,

(1) $\underline{R_B}A(x) = \inf_{u \in U}\max\{1 - \tilde{R}_B(x,u), A(u)\}$,

(2) $\overline{R_B}A(x) = \sup_{u \in U}\min\{\tilde{R}_B(x,u), A(u)\}$.

$\underline{R_B}A$ and $\overline{R_B}A$ are called the lower and upper approximation operators of $A$ on attribute subset $B$, respectively. The robust generalized fuzzy approximation operators, proposed in [11], are defined as follows.

**Definition 2.2.** Given $\alpha \in [0,1)$, let $\tilde{R}(\cdot,\cdot)$ be a fuzzy similarity relation on $U$, a robust generalized **FRS** is an ordered pair $(\underline{R_{S_\alpha}}A, \overline{R_{\sigma_\alpha}}A)$ of a fuzzy set $A$ on $U$ such that for every $x \in U$,

(1) Lower approximation operator:

$\underline{R_{S_\alpha}}A(x) = \inf_{A(u) \leq \alpha} S\left(N\left(\tilde{R}(x,u)\right), \alpha\right) \wedge \inf_{A(u) > \alpha} S\left(N\left(\tilde{R}(x,u)\right), A(u)\right);$

(2) Upper approximation operator:

$\overline{R_{\sigma_\alpha}}A(x) = \sup_{A(u) \geq N(\alpha)} \sigma(N\left(\tilde{R}(x,u)\right), N(\alpha)) \vee \sup_{A(u) < N(\alpha)} \sigma(N\left(\tilde{R}(x,u)\right), A(u)),$

where $S$ denotes an upper semi-continuous $T-$conorm, $N$ denotes an involutive negator, and $\sigma$ denotes the $T-$residuated implication based on $S$. Please refer to appendix for their definitions.

In most of the practical applications, only the decision attributes are crisp, whereas the condition attributes are continuous. Therefore, hereinafter, we mainly focus on fuzzy decision tables with crisp decision attributes.

*C. Discernibility Vector-based Rule Induction*

This subsection reviews a state-of-the-art rule induction method based on FRSs [7]. The objective of rule induction is to induce an if-then production rule from an original instance by attribute-value reduction, wherein the value reduction of each original instance as the former, and the decision label as the latter of the induced rule. The key idea of attribute-value reduction is to preserve discernibility information invariant when eliminating the redundant attribute values. Thus, the reduction of attribute values allows the fuzzy production rule to keep the main information of the original instances invariant.

In the following, we briefly review some basic concepts of value reduction [7].

**Definition 2.3 (Consistence degree).** In $FD = (U, C \cup D)$, the consistence degree of $x \in U$ is defined as $Con_{C,\alpha}^U(x) = \underline{R_{S_\alpha}}([x]_D)(x)$, where $[x]_D = \{y \in U | \tilde{R}_D(x,y) = 1\}$ consists of the instances with the same decision classes of $x$ in $U$.

Consistence degree, leveraging the discernibility information of each instance, is crucial to designing value reduction.

**Definition 2.4 (Value reduction).** In $FD = (U, C \cup D)$, if $B \subseteq C$ satisfies (1) $Con_{C,\alpha}^U(x) = Con_{B,\alpha}^U(x)$; (2) $\forall r \in B$, $Con_{C,\alpha}^U(x) \neq Con_{B-\{r\},\alpha}^U(x)$. Then, $B(x) = \{r(x): r \in B\}$ is called the value reduction of $x \in U$.

Value reduction is the minimal value subset keeping the discernable information of one instance invariant.

We denote a $n \times 1$ vector ($c_j$), called the discernibility vector of instance $x$, such that $c_j = \{a: \sigma(N(\tilde{a}(x, x_j)), Con_{C,\alpha}^U(x)) \leq \alpha\}$ for $\tilde{R}_D(x, x_j) = 0$; otherwise, $c_j = \emptyset$. Based on the discernibility vector, the value reduction algorithm is defined as follows.

---
**Algorithm 2.1.** Discernibility-vector-based Value Reduction (**DVR**)

**Input:** $FD = (U, C \cup D)$; $\alpha \in [0,1)$;
**Output:** Value reduction of $FD$: $\{reduct(x) | x \in U\}$;
**Step 1:** **For** ($i = 1$ to $|U|$) **do**
**Step 2:** $\quad V \leftarrow \{c_{ji}\}$;
**Step 3:** $\quad$ Compute $Core_\alpha(x_i) = \{a: c_{ji} = \{a\}\}$;
**Step 4:** $\quad B \leftarrow Core_\alpha(x_i), V \leftarrow V - \{v \in V | v \cap Core_\alpha(x_i) \neq \emptyset \text{ or } v = \emptyset\}$, $lef \leftarrow C - Core_\alpha(x_i)$;
**Step 5:** $\quad$ **While** ($V \neq \emptyset$), **do**
**Step 6:** $\quad\quad a^* = \arg\max_{a \in lef} |\{v \in V | a \in v\}|$;
**Step 7:** $\quad\quad B \leftarrow B \cup \{a^*\}$; $lef \leftarrow lef - \{a^*\}$;
**Step 8:** $\quad\quad V \leftarrow V - \{v \in V | v \cap \{a^*\} \neq \emptyset \text{ or } v = \emptyset\}$;
**Step 9:** $\quad$ **End while**
**Step 10:** $\quad$ Remove the superfluous attribute from $B$;
**Step 11:** $\quad reduct(x_i) \leftarrow B$;
**Step 12:** **End for**
**Step 13: Return** $\{reduct(x) | x \in U\}$.

---

From the logical point of view, in a fuzzy decision table, each original instance may be seen as a decision rule [26]. As the rule corresponding to each original instance is trivial, it is necessary to induce a generalized one by exploiting value reduction. This induced rule, whose former is composed of value reduction, provides a synthetic representation of knowledge contained in the given decision table; and then each value reduction corresponds to one induced rule. Without loss of generality, value reduction is equivalent to rule induction in **FRS**. Thus, **DVR** may be seen as a rule induction algorithm. By **DVR**, a typical rule classifier building method, named Generalized Fuzzy Rough Classifier (**GFRC**), is then designed in Algorithm A.1 (Please find it in Appendix II) [7]. Their obtained rule set works as a classifier to predict unseen instances.

## III. CVR: An alternative to value-reduction DVR

As **DVR** is space consuming, we designed its alternative, consistence degree-based attribute-value reduction (**CVR**), in this section.

First, we present a few properties of the consistence degree that are helpful in designing the **CVR**.

**Proposition 3.1.** The consistence degree of $x$ can be simplified as $Con_{C,\alpha}^U(x) = \underline{R_S}_\alpha([x]_D)(x) = \min_{y \in U \& \tilde{R}_D(x,y)=0} S(N(\tilde{R}_C(x,y)), \alpha)$.

**Proof.** By Definition 2.2, $\underline{R_S}_\alpha([x]_D)(x)$
$= \inf_{y \in U \& \tilde{R}_D(x,y)=0} S(N(\tilde{R}_C(x,y)), \alpha) \wedge \inf_{y \in U \& \tilde{R}_D(x,y)=1} S(N(\tilde{R}_C(x,y)), 1)$
$= \min_{y \in U, \tilde{R}_D(x,y)=0} S(N(\tilde{R}_C(x,y)), \alpha)$. By Definition 2.3,
$Con_{C,\alpha}^U(x) = \min_{y \in U, \tilde{R}_D(x,y)=0} S(N(\tilde{R}_C(x,y)), \alpha)$ holds. ■

When $\alpha$ is zero, the consistence degree degenerates to $Con_C^U(x) = \min_{y \in U, \tilde{R}_D(x,y)=0} N(\tilde{R}_C(x,y))$. Where $N(\tilde{R}_C(x,y))$ is the distance of $x$ from its heterogeneous instance $y$. This result shows that the consistence degree is the lower boundary at which $x$ is discerned from the heterogeneous instances in $U$ (presented in Fig. 3.1).

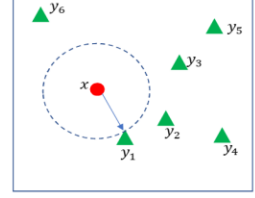

Fig. 3.1 Illustration of consistence degree on attribute set C. The dot denotes the instance $x$, and the triangles denote instances with different label from $x$. The radius of the dotted circle, i.e., the arrow, is the minimal distance from instance $x$ to the heterogeneous instances represented by triangles. The length of this radius is the consistence degree.

**Definition 3.1.** In $FD = (U, C \cup D)$, if $S(N(R_C(x,y)), \alpha) \geq Con_{C,\alpha}^U(x)$ for $\tilde{R}_D(x,u) = 0$, then $x$ and $y$ are discernible by attribute set $C$.

The monotonicity of consistence degree with gradually arriving attribute values, which is key to design an attribute-value reduction algorithm, is discussed as follows.

**Proposition 3.2 (Monotonicity of consistence degree).** Given $\alpha \in [0,1)$, if $P \subseteq Q \subseteq C$, then $Con_{P,\alpha}^U(x) \leq Con_{Q,\alpha}^U(x) \leq Con_{C,\alpha}^U(x)$.

**Proof.** By Proposition 3.1 and $\tilde{R}_P(x,y) \geq \tilde{R}_Q(x,y) \geq \tilde{R}_C(x,y)$, the result is straightforward. ■

Proposition 3.2 verifies the feasibility of designing a forward attribute-value reduction algorithm using the consistence degree.

**Definition 3.2.** Given $B \subseteq C$, $\forall a \in C - B$, the significance of $a$ in $B$ is defined as $Sig_1(a, B, x, U) = Con_{B \cup \{a\}, \alpha}^U(x) - Con_{B,\alpha}^U(x)$.

When a new attribute value is added, the increment of consistence degrees leverages the attribute-value's significance in $B$ on $x$. Then, it is feasible to design the attribute-value reduction algorithm by using this definition. We detail the proposed value reduction method through Algorithm 3.1.

---
**Algorithm 3.1.** Consistence degree-based attribute-value reduction (**CVR**)

**Input:** $FD = (U, C \cup D)$; $\alpha \in [0,1)$;
**Output:** Value reduction of $FD$: $\{reduct(x) | x \in U\}$;
**Step 1:** Calculate $Con_{C,\alpha}^U(x)$ for every $x \in U$;
**Step 2: For** every $x \in U$ **do**
**Step 3:** $\quad B \leftarrow \emptyset$, $lef \leftarrow C$;
**Step 4:** $\quad$ **While** ($Con_{B,\alpha}^U(x) < Con_{C,\alpha}^U(x)$), **do**
**Step 5:** $\quad\quad a^* = \arg\max_{a \in lef} Sig_1(a, B, x, U)$;
**Step 6:** $\quad\quad B \leftarrow B \cup \{a^*\}$; $lef \leftarrow lef - \{a^*\}$;
**Step 7:** $\quad\quad$ Update $Con_{B,\alpha}^U(x)$;
**Step 8:** $\quad$ **End while**
**Step 9:** $\quad P \leftarrow B$;
**Step 10:** $\quad$ **For** $i = 1$ to $|P|$ **do**
**Step 11:** $\quad\quad$ **If** $b_i \in B$ s.t. $Con_{B-\{b_i\},\alpha}^U(x) = Con_{C,\alpha}^U(x)$,
**Step 12:** $\quad\quad\quad B = B - \{b_i\}$;
**Step 13:** $\quad\quad$ **End if**
**Step 14:** $\quad$ **End for**
**Step 15:** $\quad reduct(x) \leftarrow B$;
**Step 16: End for**
**Step 17: Return** $\{reduct(x) | x \in U\}$.

---

**CVR** exploits the forward-addition strategy to successively add the most significant attribute-values to the candidate value reduction, until the consistence degree of $x$ reaches its maximum. It then employs a backward-deletion strategy to remove the redundant attribute-values.

Both **CVR** and **DVR** are designed based on the same idea of attribute-value reduction, i.e., keeping the discernibility information invariant. They use the discernibility vector and the consistence degree as the measure of discernibility information, respectively. At the same threshold, **CVR** and **DVR** yield equivalent value reduction as both of them meet the value reductions requirement of keeping the invariance of consistence degree invariant. Replacing **DVR** in Algorithm A.1 by **CVR**, we obtained the consistence degree-based rule classifier (i.e., **CVRC)**, which is an alternative to **GFRC**.

**CVR** is time-consuming because of the time complexity of $Con_{C,\alpha}^{U}(x)$ is $O(|U|^2|C|)$. This makes it inefficient and impractical when working on large-scale data. This calls for a mature, accelerated rule-induction algorithm that can significantly expedite fuzzy rough based rule classifier building.

## IV. A-CVR: Accelerated Rule Induction in FRSs

This study aims to accelerate fuzzy rough based rule induction considering its space and time computational limitations. In this section, we propose an accelerator of attribute-value reduction, which is equivalent to accelerating rule induction. As already mentioned, **CVR** takes much longer time to reduce the attribute values. Logically, the value reduction can be accelerated by compacting its search space. In the following we propose a pair of new categories, Key Set and Discernible Set, which are composed of key elements for value reduction.

### A. Discernible Set and Key Set

#### 1) Discernible set and Key set

Consistence degree is a boundary, keeping which invariant, the redundant values can be detected and then reduced. However, the computation of this degree has to explore the entire space. Some interesting properties of the consistence degree are discussed and presented as follows.

**Theorem 4.1**. In $FD = (U, C \cup D)$, given $\alpha \in [0,1)$, for $B \subseteq C$ and $x, u \in U$ s.t. $\tilde{R}_D(x,u) = 0$, the following statements hold.
(1) If $S(N(\tilde{R}_B(x,u)), \alpha) \geq Con_{C,\alpha}^{U}(x)$, then $x$ and $u$ are discernible by attribute set $B$;
(2) $Con_{C,\alpha}^{U}(x) = \min_{y \in H(x)} S(N(\tilde{R}_C(x,y)), \alpha)$,
where $H(x) = \{u \in U | S(N(\tilde{R}_B(x,u)), \alpha) < Con_{C,\alpha}^{U}(x) \text{ and } \tilde{R}_D(x,u) = 0\}$.
**Proof.**
(1) $B \subseteq C \Rightarrow \tilde{R}_B(x,u) \geq \tilde{R}_C(x,u) \Rightarrow N(\tilde{R}_B(x,u)) \leq N(\tilde{R}_C(x,u))$
$\Rightarrow S(N(\tilde{R}_B(x,u)), \alpha) \leq S(N(\tilde{R}_C(x,u)), \alpha)$
$\Rightarrow S(N(\tilde{R}_C(x,u)), \alpha) \geq Con_{C,\alpha}^{U}(x)$ as $S(N(\tilde{R}_B(x,u)), \alpha) \geq Con_{C,\alpha}^{U}(x)$.

Thus, $x$ and $u$ are discernible by attribute set $B$ according to **Definition 3.1**.
(2) By **Proposition 3.1**, $Con_{C,\alpha}^{U}(x) = \min_{y \in U, R_D(x,y)=0} S(N(\tilde{R}_C(x,y)), \alpha)$. Then
$Con_{C,\alpha}^{U}(x) = \min_{y \in H(x)} S(N(\tilde{R}_C(x,y)), \alpha) \wedge \min_{y \in H'(x)} S(N(\tilde{R}_C(x,y)), \alpha)$,
where $H'(x) = \{u \in U | S(N(\tilde{R}_B(x,u)), \alpha) \geq Con_{C,\alpha}^{U}(x) \text{ and } \tilde{R}_D(x,u) = 0\}$ and $H'(x) \cup H(x) = \{u \in U | \tilde{R}_D(x,u) = 0\}$.
By (1), we have $\min_{y \in H'(x)} S(N(\tilde{R}_C(x,y)), \alpha) \geq Con_{C,\alpha}^{U}(x)$.
Thus, $Con_{C,\alpha}^{U}(x) = \min_{y \in H(x)} S(N(\tilde{R}_C(x,y)), \alpha)$. ∎

**Fig.4.1** visualizes the results of **Theorem 4.1**. From **Fig.4.1**, we observe the following facts.
- The distance between instances widens as the attributes increase. This fact has been verified in the proof of Theorem 4.1(1).
- The triangles with black borders always stay outside the dotted circle on attribute subset $B$ or $C$. By Theorem 4.1(1), it is concluded that the triangles with black borders are discernible from $x$ by attribute subset $B$.
- Fig. 4.1 illustrates that the computation of the consistence degree is not related to the triangles with the black borders, which is the fact revealed by Theorem 4.1 (2).

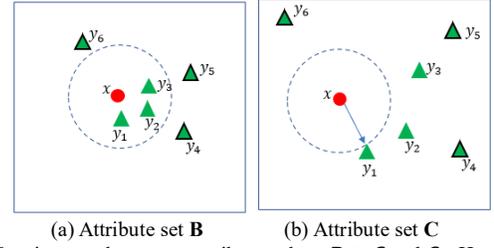

(a) Attribute set **B**     (b) Attribute set **C**

Fig. 4.1. Consistence degree on attribute subset $B \subseteq C$ and $C$. Here, the red dot denotes instance $x$, and the triangles denote instances with different labels from $x$. The radius of the dotted circle, i.e., the arrow, is the minimal distance from instance $x$ to the heterogeneous instances represented by a triangle.

By Theorem 4.1, it is practicable to quickly update the consistence degree; avoid recomputing it on the whole universe. We form a special set of such instances that were valid for updating the consistence degree.

**Definition 4.1(Discernible Set and Key set).** In $FD = (U, C \cup D)$, given $B \subseteq C$ and $\alpha \in [0,1)$, **Discernible set** of $x \in U$ with respect to $B$ is defined as $Dis_B(x) = \{u \in U | S(N(\tilde{R}_B(x,u)), \alpha) \geq Con_{C,\alpha}^{U}(x) \text{ and } \tilde{R}_D(x,u) = 0\}$; **Key set** of $x \in U$ with respect to $B$ is defined as $KEY_B(x) = \{y \in U - Dis_B(x) | \tilde{R}_D(x,y) = 0\}$.

Discernible Set of $x$ composes of its heterogeneous instances that can be discerned from $x$ on $B$; Key Set of $x$ composes of the remainder heterogeneous instances, that were undiscerned from $x$ by $B$. Discernible Set and Key Set are illustrated in **Fig. 4.1**. In the left-hand side illustration of **Fig. 4.1** the encircled triangle instances constitute Key Set of $x$ on $B$, whereas those outside the circle constitute Discernible Set of $x$ on $B$. Some of their properties are presented below.

#### 2) Properties of Discernible set and Key set

Discernible set and Key set have the following properties.

**Proposition 4.1 (Monotonicity of Discernible Set and Key set).** In $FD = (U, C \cup D)$, given $B_i \subseteq C$, $x \in U$ and $\alpha \in [0,1)$, let $B_1 \subseteq B_2 \subseteq \cdots \subseteq B_n$, we have
(1) $Dis_{B_1}(x) \subseteq Dis_{B_2}(x) \subseteq \cdots \subseteq Dis_{B_n}(x)$;
(2) $|Dis_{B_1}(x)| \leq |Dis_{B_2}(x)| \leq \cdots \leq |Dis_{B_n}(x)|$;
(3) If $y \in U - Dis_{B_i}(x)$ and $\tilde{R}_D(x,y) = 0$, then $S(N(\tilde{R}_{B_i}(x,u)), \alpha) < Con_{C,\alpha}^{U}(x)$;
(4) $Dis_{B_i}(x) \cup KEY_{B_i} = \{y \in U | \tilde{R}_D(x,y) = 0\}$;
(5) $KEY_{B_1}(x) \supseteq KEY_{B_2}(x) \supseteq \cdots \supseteq KEY_{B_n}(x)$;
(6) $|KEY_{B_1}(x)| \geq |KEY_{B_2}(x)| \geq \cdots \geq |KEY_{B_n}(x)|$.
**Proof.**
(1) $B_i \subseteq B_j \Rightarrow \tilde{R}_{B_i}(x,u) \geq \tilde{R}_{B_j}(x,u) \Rightarrow N(\tilde{R}_{B_i}(x,u)) \leq N(\tilde{R}_{B_j}(x,u))$
$\Rightarrow S(N(\tilde{R}_{B_i}(x,u)), \alpha) \leq S(N(\tilde{R}_{B_j}(x,u)), \alpha)$
$\Rightarrow Dis_{B_i}(x) \subseteq Dis_{B_j}(x)$ by Definition 4.1.
(2) The result is straightforward by the result in (1).
(3) $y \in U - Dis_{B_i}(x)$ and $\tilde{R}_D(x,y) = 0 \Rightarrow S(N(\tilde{R}_{B_i}(x,u)), \alpha) < Con_{C,\alpha}^{U}(x)$. (4)(5) and (6) are straightforward. ∎

Proposition 4.1(1) & (2) prove that $B_j$ has a stronger power of discernibility than $B_i$ as the number of attributes increases. In addition, the size of Discernible Set $Dis_{B_i}(x)$ increases monotonically as $B_i$. Furthermore, Proposition 4.1(3) & (4) present that Discernible Set and Key Set complement each other in the set of heterogeneous instances of $x$. Finally, according to Proposition 4.1(5) & (6), Key Set becomes more and more compact as the attributes successively increases. Based on the result of Proposition 4.1, some of the deeper properties are revealed below.

**Proposition 4.2.** In $FD = (U, C \cup D)$, given $B \subseteq P \subseteq C$, $x \in U$ and $\alpha \in [0,1)$, the following statements hold.

(1) $Con_{B,\alpha}^U(x) = \min_{y \in KEY_B(x)} S(N(\tilde{R}_B(x,y)), \alpha) \wedge \min_{y \in Dis_B(x)} S(N(\tilde{R}_B(x,y)), \alpha)$;

(2) $\min_{y \in Dis_B(x)} S(N(\tilde{R}_P(x,y)), \alpha) \geq Con_{C,\alpha}^U(x)$;

(3) $Con_{B,\alpha}^U(x) < Con_{C,\alpha}^U(x) \Leftrightarrow KEY_B(x) \neq \emptyset \Leftrightarrow Con_{B,\alpha}^U(x) = \min_{y \in KEY_B(x)} S(N(\tilde{R}_B(x,y)), \alpha)$;

(4) $KEY_B(x) = \emptyset \Leftrightarrow Con_{C,\alpha}^U(x) = Con_{B,\alpha}^U(x)$;

(5) $KEY_B(x) = \emptyset \Rightarrow KEY_P(x) = \emptyset$.

**Proof.**

(1) $Con_{B,\alpha}^U(x) = \min_{y \in U \text{ and } R_D(x,y)=0} S(N(\tilde{R}_B(x,y)), \alpha)$
$\Rightarrow Con_{B,\alpha}^U(x) = \min_{y \in U - Dis_B(x) \text{ and } R_D(x,y)=0} S(N(\tilde{R}_B(x,y)), \alpha) \wedge \min_{y \in Dis_B(x)} S(N(\tilde{R}_B(x,y)), \alpha)$
$\Rightarrow Con_{B,\alpha}^U(x) = \min_{y \in KEY_B(x)} S(N(\tilde{R}_B(x,y)), \alpha) \wedge \min_{y \in Dis_B(x)} S(N(\tilde{R}_B(x,y)), \alpha)$.

(2) $\forall y \in Dis_B(x)$,
$\Rightarrow S(N(\tilde{R}_B(x,u)), \alpha) \geq Con_{C,\alpha}^U(x)$, $\forall y \in Dis_B(x)$
$\Rightarrow S(N(\tilde{R}_P(x,u)), \alpha) \geq S(N(\tilde{R}_B(x,u)), \alpha) \geq Con_{C,\alpha}^U(x)$ as $B \subseteq P$, $\forall y \in Dis_B(x)$
$\Rightarrow \min_{y \in Dis_B(x)} S(N(\tilde{R}_P(x,y)), \alpha) \geq Con_{C,\alpha}^U(x)$.

(3) $Con_{B,\alpha}^U(x) < Con_{C,\alpha}^U(x)$
$\Leftrightarrow$ There exists one instance $\in U$, which is heterogeneous instance of $x$, i.e., $\tilde{R}_D(x,y) = 0$, satisfying $S(N(\tilde{R}_B(x,y)), \alpha) = Con_{B,\alpha}^U(x) < Con_{C,\alpha}^U(x)$
$\Leftrightarrow KEY_B(x) \neq \emptyset$ by the definition of $KEY_B(x)$
$\Leftrightarrow Con_{B,\alpha}^U(x) = \min_{y \in KEY_B} S(N(\tilde{R}_B(x,y)), \alpha)$ as $Con_{B,\alpha}^U(x) = \min_{y \in KEY_B} S(N(\tilde{R}_B(x,y)), \alpha) \wedge \min_{y \in Dis_B(x)} S(N(\tilde{R}_B(x,y)), \alpha)$.

(4) $KEY_B(x) = \emptyset$
$\Leftrightarrow \forall y \in U$ s.t. $\tilde{R}_D(x,y) = 0$, $S(N(\tilde{R}_B(x,y)), \alpha) \geq Con_{C,\alpha}^U(x)$
$\Leftrightarrow Con_{B,\alpha}^U(x) = \min_{y \in Dis_B} S(N(\tilde{R}_B(x,y)), \alpha) = Con_{C,\alpha}^U(x)$.

(5) $KEY_B(x) = \emptyset$
$\Rightarrow \forall y \in U$ s.t. $\tilde{R}_D(x,y) = 0$, $S(N(\tilde{R}_B(x,y)), \alpha) \geq Con_{C,\alpha}^U(x)$
$\Rightarrow \forall y \in U$ s.t. $\tilde{R}_D(x,y) = 0$, $S(N(\tilde{R}_P(x,y)), \alpha) \geq Con_{C,\alpha}^U(x)$ since $S(N(\tilde{R}_P(x,y)), \alpha) \geq S(N(\tilde{R}_P(x,y)), \alpha)$
$\Rightarrow KEY_P(x) = \emptyset$. ∎

Proposition 4.2 describes the properties of the consistence degree on Discernible Set and Key Set. Proposition 4.3 describes the characteristics of the consistence degree in some special cases of Key Set.

**Proposition 4.3.** In $FD = (U, C \cup D)$, given $B \subseteq C$, $x \in U$ and $\alpha \in [0,1)$, $Dis_B(x)$ is the discernible set of $B$ on $x$, we have

(1) If $KEY_B(x) \neq \emptyset$, then $Con_{B,\alpha}^U(x) = Con_{B,\alpha}^{U-Dis_B(x)}(x)$;

(2) If $KEY_B(x) = \emptyset$, then $Con_{B,\alpha}^U(x) = Con_{C,\alpha}^U(x)$.

**Proof.**

(1) $KEY_B(x) = \{y \in U - Dis_B(x) | \tilde{R}_D(x,y) = 0\} \neq \emptyset$
$\Rightarrow Con_{B,\alpha}^U(x) < Con_{C,\alpha}^U(x)$
$\Rightarrow Con_{B,\alpha}^U(x) = \min_{y \in KEY_B(x)} S(N(\tilde{R}_B(x,y)), \alpha) = Con_{B,\alpha}^{U-Dis_B(x)}(x)$.

(2) $KEY_B(x) = \emptyset \Rightarrow Con_{B,\alpha}^U(x) = Con_{C,\alpha}^U(x)$. ∎

Based on these properties of Discernible Set and Key Set, the significance degree of the attribute value on $x$ is designed in a new way.

**Definition 4.2.** In $FD = (U, C \cup D)$, given $B \subseteq C$ and $\alpha \in [0,1)$, for $x \in U$, if $Dis_B(x)$ and $KEY_B(x)$ are Discernible Set and Key Set of $B$ on $x$ (i.e., $Dis_B(x) = \{U | \tilde{R}_D(x,y) = 0\} - KEY_B(x)$), respectively, then $\forall a \in C - B$,

$Sig_2(a, B, x, KEY_B(x))$
$= \begin{cases} Con_{B \cup \{a\}, \alpha}^{U-Dis_B(x)}(x) - Con_{B,\alpha}^{U-Dis_B(x)}(x), & KEY_B(x) \neq \emptyset, \text{ and } KEY_{B \cup \{a\}}(x) \neq \emptyset \\ Con_{C,\alpha}^U(x) - Con_{B,\alpha}^{U-Dis_B(x)}(x), & KEY_B(x) \neq \emptyset, \text{ but } KEY_{B \cup \{a\}}(x) = \emptyset \\ 0, & KEY_B(x) = \emptyset \end{cases}$

is called the **relative significance degree** of $a$ in $B$ on $x$ with respect to $D$.

According to Definition 4.2, the computation of the **relative significance degree** is not based on the whole universe, but on Key Set. Then, by using **relative significance degree**, it is feasible to design the value-reduction algorithm without recomputing on the whole universe.

*3) Main theorem of Key Set and Discernible Set*

In the design of attribute-value reduction algorithm, we found that Key Set and Discernible Set yielded a more interesting result, the rank preservation property of the relative significance degree.

**Theorem 4.2 (Rank Preservation Property).** In $D = (U, C \cup D)$, given $B \subseteq C$ and $\alpha \in [0,1)$. $\forall a, b \in C - B$, if $Sig_1(a, B, x, U) \geq Sig_1(b, B, x, U)$, then $Sig_2(a, B, x, KEY_B(x)) \geq Sig_2(b, B, x, KEY_B(x))$.

**Proof.**
$Sig_1(a, B, x, U) \geq Sig_1(b, B, x, U)$
$\Rightarrow Con_{B \cup \{a\}, \alpha}^U(x) - Con_{B,\alpha}^U(x) \geq Con_{B \cup \{b\}, \alpha}^U(x) - Con_{B,\alpha}^U(x)$.
$\Rightarrow Con_{B \cup \{a\}, \alpha}^U(x) \geq Con_{B \cup \{b\}, \alpha}^U(x)$.

In the case of $KEY_{B \cup \{a\}}(x) \neq \emptyset$ and $KEY_{B \cup \{b\}}(x) \neq \emptyset$,
$Con_{B \cup \{a\}, \alpha}^U(x) \geq Con_{B \cup \{b\}, \alpha}^U(x)$
$\Rightarrow Con_{B \cup \{a\}, \alpha}^{U-Dis_B(x)}(x) \geq Con_{B \cup \{b\}, \alpha}^{U-Dis_B(x)}(x)$
$\Rightarrow Con_{B \cup \{a\}, \alpha}^{U-Dis_B(x)}(x) - Con_{B,\alpha}^{U-Dis_B(x)}(x) \geq Con_{B \cup \{b\}, \alpha}^{U-Dis_B(x)}(x) - Con_{B,\alpha}^{U-Dis_B(x)}(x)$
$\Rightarrow Sig_2(a, B, x, KEY_B(x)) \geq Sig_2(b, B, x, KEY_B(x))$.

In the case of $KEY_{B \cup \{a\}}(x) \neq \emptyset$ and $KEY_{B \cup \{b\}}(x) = \emptyset$,
$Con_{B \cup \{a\}, \alpha}^U(x) \geq Con_{B \cup \{b\}, \alpha}^U(x)$
$\Rightarrow Con_{B \cup \{a\}, \alpha}^{U-Dis_B(x)}(x) - Con_{B,\alpha}^{U-Dis_B(x)} \geq Con_{C,\alpha}^U(x) - Con_{B,\alpha}^{U-Dis_B(x)}$
$\Rightarrow Sig_2(a, B, x, KEY_B(x)) \geq Sig_2(b, B, x, KEY_B(x))$.

In the case of $KEY_{B \cup \{a\}}(x) = \emptyset$ and $KEY_{B \cup \{b\}}(x) = \emptyset$,
$Con_{B \cup \{a\}, \alpha}^U(x) \geq Con_{B \cup \{b\}, \alpha}^U(x)$
$\Rightarrow Con_{C,\alpha}^U(x) - Con_{B,\alpha}^{U-Dis_B(x)} = Con_{C,\alpha}^U(x) - Con_{B,\alpha}^{U-Dis_B(x)}$
$\Rightarrow Sig_2(a, B, x, KEY_B(x)) = Sig_2(b, B, x, KEY_B(x))$.

In all cases, $Sig_1(a, B, x, U) \geq Sig_1(b, B, x, U)$
$\Rightarrow Sig_2(a, B, x, KEY_B(x)) \geq Sig_2(b, B, x, KEY_B(x))$. ∎

Theorem 4.2 is an important result of Key Set and Discernible set, which verifies that the rank of the significance degrees computed on the whole universe is consistent with that computed on Key Set. Thus, it is sufficient to find the value reduction by only updating the consistence degree on Key Set. This mechanism can then be used to improve the computational efficiency of a heuristic value-reduction algorithm, while producing the same result.

*B. Attribute-value Reduction Accelerator*

In this subsection, we propose an accelerated way to update the consistence degree based on Discernible Set and Key Set.

**Theorem 4.3.** In $FD = (U, C \cup D)$, $B \subseteq C$ and $\alpha \in [0,1)$, for $x \in U$, the following statements hold.

(1) If $Dis_B(x) = \{y \in U | \tilde{R}_D(x,y) = 0\}$, then $B(x) = \{r(x) : r \in B\}$ contains the attribute-value reduction of $x \in U$.

(2) If $KEY_B(x) = \emptyset$, then $B(x) = \{r(x) : r \in B\}$ contains the attribute-value reduction of $x \in U$.

**Proof.**

(1) $Dis_B(x) = \{y \in U | \tilde{R}_D(x,u) = 0\}$
$\Rightarrow \forall y \in U$ s.t. $\tilde{R}_D(x,y) = 0$, $S(N(\tilde{R}_B(x,y)), \alpha) \geq Con_{C,\alpha}^U(x)$

$\Rightarrow \min_{y \in U, R_D(x,y)=0} S(N(\tilde{R}_B(x,y)), \alpha) \geq Con_{C,\alpha}^U(x)$

$\Rightarrow Con_{B,\alpha}^U(x) \geq Con_{C,\alpha}^U(x)$

$\Rightarrow B(x)$ contains the attribute-value reduction of $x \in U$.

(2) As $Dis_B(x) = \{y \in U | \tilde{R}_D(x,u) = 0\}$ is equivalent to $KEY_B(x) = \emptyset$, it is easy to get the result. ∎

**Theorem 4.3** shows that, when the size of the discernible set $Dis_B(x)$ is large enough to contain all the heterogeneous instances of $x$ or the key set $KEY_B(x)$ is small enough to be empty, then $B(x)$ contains one attribute-value reduction of $x \in U$. By **Theorem 4.3**, we design an accelerator of **CVR** as shown in **Algorithm 4.1**.

**Algorithm 4.1.** Consistence degree-based attribute-value reduction accelerator (**A-CVR**)

**Input:** $FD = (U, C \cup D); \alpha \in [0,1)$;
**Output:** Value reduction of $FD$: $\{reduct(x) | x \in U\}$;
**Step 1: For** each $x \in U$ **do**
**Step 2:**    $i = 1, U_1 = U; B \leftarrow \emptyset, lef \leftarrow C$;
**Step 3:**    Compute $Con_{C,\alpha}^U(x); Dis_B(x); KEY_B(x)$;
**Step 4:**    **While** ($KEY_B(x) \neq \emptyset$), **do**
**Step 5:**        $a^* = \arg\max_{a \in lef} Sig_2(a, B, x, KEY_B(x))$;
**Step 6:**        $B \leftarrow B \cup \{a^*\}, lef \leftarrow lef - \{a^*\}$;
**Step 7:**        Update $Dis_B(x); KEY_B(x); U_{i+1} \leftarrow U_i - Dis_B(x); i \leftarrow i + 1$;
**Step 8:**    **End while**
**Step 9:**    $P \leftarrow B$;
**Step 10:**    **For** $i = 1$ to $|P|$ **do**
**Step 11:**        **If** $b_i \in B$ s.t. $Con_{B-\{b_i\},\alpha}^U(x) = Con_{C,\alpha}^U(x)$,
**Step 12:**            $B = B - \{b_i\}$;
**Step 13:**        **End if**
**Step 14:**    **End for**
**Step 15:**    $reduct(x) \leftarrow B$;
**Step 16: End for**
**Step 17: Return** $\{reduct(x) | x \in U\}$.

**A-CVR** is the accelerated version of **CVR**. Unlike the search space that remains fixed in **CVR**, Key Set becomes smaller with successively added attribute values in **A-CVR**. Thus, the redundant computation of value reduction is reduced in **A-CVR**. **A-CVR** and **CVR** are compared in detail as follows.

- **Theorem 4.2** ensures that Step 5 in **A-CVR** is equivalent to Step 5 in **CVR**; and they can choose the same significant attribute value in each iteration.
- **Theorem 4.3** ensures that Step 4 in **A-CVR** is equivalent to Step 4 in **CVR**. Accordingly, their stop criteria are equivalent.
- By both **Theorems 4.2** and **4.3**, **A-CVR** is equivalent to **CVR**; they are verified to obtain the same attribute-value reduction.
- Except Steps 5 and 7, **A-CVR** and **CVR** share the same time complexity. In **A-CVR**, the time complexity of Step 5 is $O(\sum_{i=1}^{|C|}(|C| + 1 - i)|KEY(B)|)$, while it is $O(\sum_{i=1}^{|C|}(|C| + 1 - i)|U|)$ in **CVR**. The time complexity of Step 7 in **A-CVR** is $O(|U_i|)$, while it is $O(|U|)$ in **CVR**. As $KEY(B)$ and $U_i$ become smaller and smaller during the iteration, it is obvious that **A-CVRA** save more time compared to **CVR**.

Theorems 4.2 & 4.3 ensure that **A-CVR** achieves the same value-reduction as **CVR** at a smaller computational cost. As value reduction is equivalent to rule induction, we refer to the accelerated rule induction as **A-CVR** hereinafter.

## V. ACCELERATED RULE BASED CLASSIFIERS

In this section, we present an accelerated classifier-building algorithm and its updated version based on accelerated rule induction **A-CVR**.

### A. Accelerated Rule-Classifier Building

In this subsection, we first employ **A-CVR** to accelerate rule-classifier building. The rule extraction strategy of **GFRC**, i.e., forward adding strategy, is then adopted to find a near-minimal rule set. Then, the accelerated version of **CVRC**, denoted by **A-CVRC**, is obtained (presented in **Algorithm 5.1**).

Similarly, when rule induction of **A-CVR** coordinates with the rule extraction strategy of **LEM2** or **VC-DomLE** (presented in Appendix II), their accelerated counterparts were then obtained, denoted by **A-LEM2** and **A-VC-DomLE**, respectively.

**Algorithm 5.1.** Accelerated **CVRC** (**A-CVRC**)

**Input:** $FD = (U, C \cup D); \alpha \in [0,1)$;
**Output**: Near-minimal rule set: $minimal\_rule\_set$;
**Step 1: Calculate** the induced rule $reduct(x)$ of every original decision rule by Algorithm 4.1, **A-CVR**;
**Step 2: Add** all $reduct(x)$ into $all\_rules$, $minimal\_rule\_set \leftarrow \emptyset$;
**Step 3: Calculate** the cover degree of every rule in $all\_rules$;
**Step 4: Extract** $minimal\_rule\_set$ from $all\_rules$ by using the rule extraction strategy of **GFRC**;
**Step 5: Return** $minimal\_rule\_set$.

### B. Updated Version: Batch Sample-based Rule Classifier

To handle large-scale datasets, we accelerate rule extraction by the Batch Sample scheme. Inspired by the idea of mini-batch in deep learning, that accelerates the training process with fewer losses, we compute the value-reduction accelerator (i.e., **A-CVR**) on a batch of instances rather than on all instances. Based on this idea, Algorithm 5.2 is then designed. Notably, when a batch is as large as the whole universe, **A-BSRC** and **A-CVRC** are essentially the same.

**Algorithm 5.2.** Batch Sample Rule Classifier (**A-BSRC**)

**Input:** $FD = (U, C \cup D); \alpha \in [0,1)$; The percent of batch sample; The cover degree threshold: $\delta \in [0, |U|]$;
**Output**: Near-minimal rule set: $\Omega$;
**Step 1: Let** the pool storing the candidate induced rule be empty, i.e., $red = \emptyset$; the pool to storing the instances be full, i.e., $T = U$;
**Step 2: Repeat**
**Step 3:**    Sample a batch $\Delta T$ from $T$, where $|\Delta T| = \lceil \beta \times |T| \rceil$;
**Step 4:**    Compute $\Delta red = \{reduct(x) = \emptyset | x \in \Delta T\}$ by **A-CVR**;
**Step 5:**    $red = red \cup \Delta red, T = T - \Delta T$;
**Step 6:**    $\forall r \in red$, compute the covering degree on $red$ and $T$;
**Step 7:**    **While** $red \neq \emptyset$ and the maximal covering degree of rule in $red$ exceeds $\delta$, **do**
**Step 8:**        Add $r^* \in red$ with the maximum covering degree into $\Omega$;
**Step 9:**        Delete rules covered by $r^*$ from $red$;
**Step 10:**        Delete instances covered by $r^*$ from $T$;
**Step 11:**        Update the covering degree of every rule in $red$;
**Step 12:**    **End while**
**Step 13: Until** $T = \emptyset$;
**Step 14: Return** $\Omega$.

## VI. NUMERICAL EXPERIMENTS

In this section, we evaluated the accelerator **A-CVR** in terms of time efficiency and classification performance, respectively. To demonstrate the efficiency of our proposed methods, we compared the computational time of our accelerated framework with the unaccelerated one. Furthermore, to demonstrate the classification performance of the proposed methods, we compared our accelerated classifier with some rough-based classifiers and some known explainable classifiers (not related to rough sets).

### A. Experimental Setup

The experiments are set up as follows.

**Datasets:** We conducted numerical experiments on a series of UCI and KEEL datasets. Following the rational of data selection in [14,24], some datasets, that differ considerably regarding number of instances, attributes and classes, were selected (as presented in Table 6.1). These datasets helped comprehensively analyze the algorithm's performance in terms of size, dimensionality and categories. They served as a good test bed for a comprehensive evaluation.

TABLE 6.1
THE DESCRIPTION OF THE SELECTED DATASETS[1].

|   | Datasets | Number of Attributes | Number of Instances | Number of Classes |
|---|---|---|---|---|
| 1 | Iono | 34 | 351 | 2 |
| 2 | Libras | 90 | 540 | 15 |
| 3 | QSAR | 41 | 1,055 | 2 |
| 4 | Cont | 9 | 1,473 | 3 |
| 5 | Segm | 19 | 2,310 | 7 |
| 6 | Spam | 57 | 4,601 | 2 |
| 7 | Texture | 40 | 5,500 | 11 |
| 8 | Optdigits | 64 | 5,620 | 10 |
| 9 | Park | 21 | 5,875 | 2 |
| 10 | Sat | 36 | 6,435 | 6 |
| 11 | Musk2 | 166 | 6,598 | 2 |
| 12 | Thyroid | 21 | 7,200 | 3 |
| 13 | Ring | 20 | 7,400 | 2 |
| 14 | coil2000 | 85 | 9,822 | 2 |
| 15 | Crowd | 28 | 10,845 | 6 |
| 16 | Pendigits | 16 | 10,992 | 10 |
| 17 | Nursery | 8 | 13,576 | 5 |
| 18 | Eeg | 14 | 14,980 | 2 |
| 19 | Shuttle | 9 | 58,000 | 7 |
| 20 | Sensorless | 48 | 58,509 | 11 |

Each condition attribute of these datasets was normalized into the interval [0, 1] with MinMaxScaler. Following [7], we selected $S(x,y) = min\{1, x + y\}$ as the $T$-conorm and $N(x) = 1 - x$ as a negator to construct the lower approximation operator. We considered the Łukasiewicz $T$-norm, $T_L(a,b) = \max\{0, a + b - 1\}, a, b \in [0,1]$, as a special case of the triangular norm $T$. Then, the similarity degree satisfying $T_L$ can be calculated as $\tilde{R}_a(x,y) =$ $1 - (\max(a(x), a(y)) - \min(a(x), a(y)))$, where $a(x), a(y) \in [0,1]$ represent the attribute values of instances $x$ and $y$ on $a$, respectively. For details, the reader is referred to [6,10].

**Parameter setting**: For **CVRC**, **A-CVRC**, and **GFRC**, the parameter employed was set as $\alpha = 0$, as the inconsistent instances (same attribute-value but different class) were removed from 'Cont' and 'Spam' in advance. For **A-BSRC**, the parameters employed were set as $\alpha = 0, \beta = 0.01, \delta = 0$. A five-fold cross-validation was performed on each dataset, and the average prediction accuracies/execution time and standard deviations were recorded. In addition, the outcomes of pairwise t-test at a significance level of 0.05 for all the experiments were recorded to investigate the algorithm—**A-CVR** or **A-BSRC**—that was significantly superior/inferior (win/loss) to the state-of-the-art algorithms.

**Environments:** All experiments detailed in this section were conducted on a computer with Windows 10 Professional, Intel(R) Xeon(R) W-2148 CPU@3.70GHz and 128GB memory. The programming language is Python 3.6.

### B. Time Efficiency Analysis: different Rule-induction methods

In this subsection, we compared the time efficiency of three classifier-building algorithms, **A-CVRC**, **CVRC** and **GFRC**. These three rule-based classifiers adopted **CVR**, **DVR** and **A-CVR**, respectively, to induce rules; but they adopted the same search strategies of rule extraction. Accordingly, it is reasonable and fair to compare the time efficiencies of different rule induction methods by exploiting these classifiers.

In Fig. 6.1, we illustrated the running time of **A-CVRC**, **CVRC** and **GFRC** on twelve datasets selected from Table 6.1.

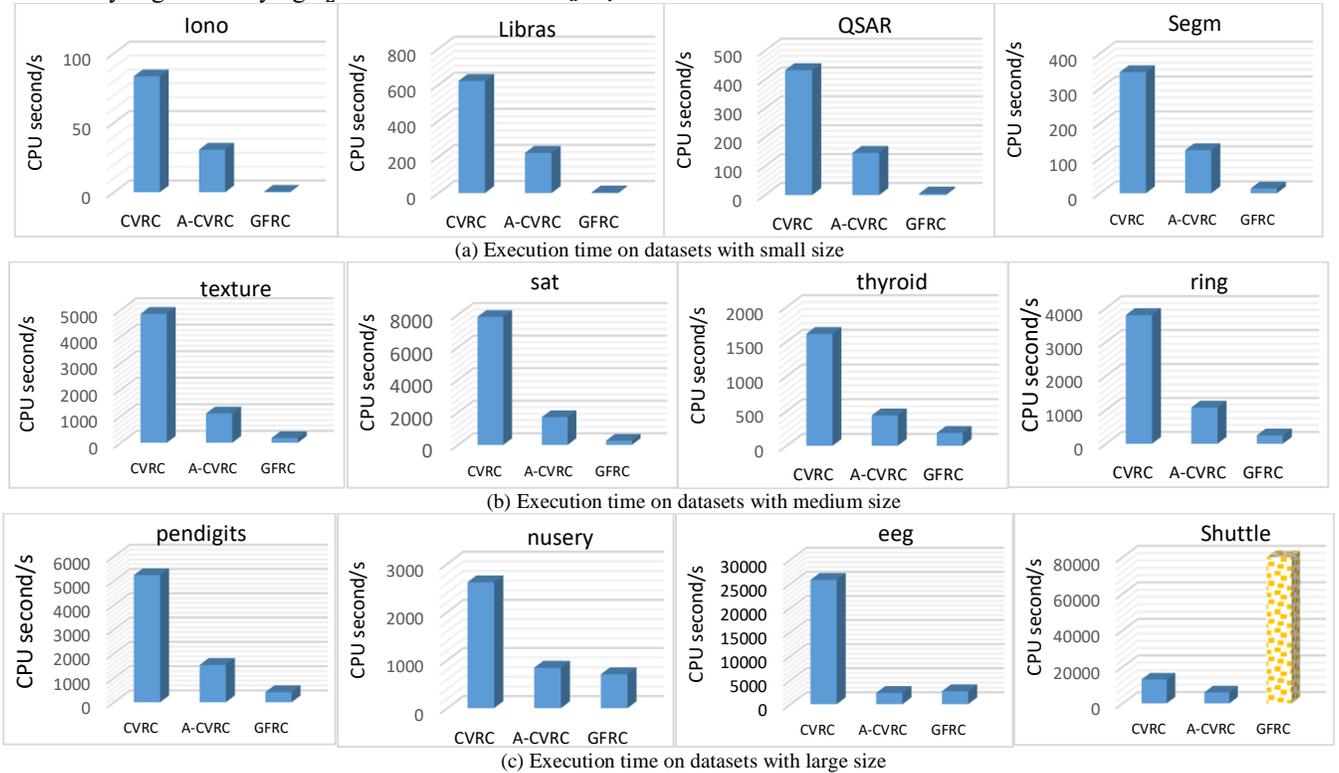

(a) Execution time on datasets with small size

(b) Execution time on datasets with medium size

(c) Execution time on datasets with large size

Fig. 6.1 The efficiency comparison among three rough-based rule classifiers: A-CVRC, CVRC, GFRC. In each sub-figure, the x,y-coordinate pertain to dataset size, and the running time(/s), respectively. The yellow dotted pillar indicates that GFRC cannot work on the corresponding dataset.

---

[1] The used datasets and the codes of the proposed methods are released in https://github.com/RUC-DWBI-ML/A-CVRC.

The following facts are derived from Fig. 6.1.
- In every subgraph except 'Shuttle', the pillar of **CVRC** is the highest one. For example, the running time of **CVRC** is obviously more than 4 times of that of **A-CVRC** on 'eeg', 'sat' and 'texture'. Moreover, the running time of **CVRC** exceeds 10 times of that of **GFRC** on most selected datasets. These charts show that **CVRC** is the most time-consuming algorithm and it is because the time complexity of **CVR** is the square of the number of instances on the whole search space.
- In every subgraph except 'Shuttle', the pillar of **GFRC** is the lowest one. For example, the running time of **GFRC** is far less than one thirtieth of that of **CVRC** on 'Iono', 'Libras' and 'QSAR'. This shows that **GFRC** is the fastest algorithm on the small-size datasets. However, it is also observed that **GFRC** fails to work on 'shuttle', indicated by the yellow dotted pillar. This is because the space complexity of **DVR** is the square of the number of instances; it is highly space consuming and then **GFRC** is impractical on large-size datasets.
- In every subgraph, we observe that the pillars of **A-CVRC** is always remarkably lower than those of **CVRC**, which shows that **A-CVRC** is always faster than its unaccelerated counterpart, **CVRC**. This empirically demonstrates the efficiency of the proposed accelerated algorithm **A-CVR**.
- It is noted that on subgraph 'eeg' with large size of instances, the pillar of **GFRC** is observably higher than that of **A-CVRC**. Additionally, **GFRC** fails to work on 'Shuttle', indicated by the yellow dotted pillar. These show that on large-size datasets, **A-CVRC** is workable and faster than **GFRC**.

These observations and facts show that **A-CVR** is an efficient and practical method due to the compacted search space, Key Set.

*C. Time Efficiency Analysis: Accelerated and Unaccelerated*

This subsection evaluated the time efficiencies of the proposed accelerated methods followed by different rule-extraction strategies. Specifically, we compared four pairs of accelerated and unaccelerated algorithms, i.e., **CVRC** and **A-CVRC**, **BSRC** and **A-BSRC**, **LEM2** and **A-LEM2**, and **VC-DomLE** and **A-VC-DomLE**. Each pair of algorithms adopted the same rule extraction strategy, but different (accelerated and unaccelerated) rule induction methods. Just as stated in Theorems 4.2 and 4.3, the proposed accelerator **A-CVR** can achieve the same induced rule as its unaccelerated counterpart, **CVR**. Consequently, each pair of algorithms achieves the same induced rule set; it is no need to compare their classification accuracy.

We employed eight datasets selected from Table 6.1 to verify the efficiency of the accelerated method. To distinguish the computational time, we divided each dataset into ten subgroups of equal size by the splitting method proposed in [14]. Specifically, eight selected datasets are split into 10 subgroups by row. One subgroup is regarded as the original data, the other subgroups are added into the original data successively. These constructed data in such ways can be used to evaluate the time taken by each pair of algorithms.

Tables 6.2-6.5 display the number of selected rules and computational time. Figs. 6.2-6.5 describe the more detailed time trendline of each pair of algorithms with respect to dataset size. We derive the following facts from the figures and the tables:
- From Figs.6.2-6.5 we observe that each blue trendline is always higher than its corresponding orange one. This demonstrates that the running time of accelerated algorithm (i.e., **A-CVRC**, **A-BSRC**, **A-LEM2** or **A-VC-DomLE**) is always notably higher than that of its unaccelerated counterpart (i.e., **CVRC**, **BSRC**, **LEM2** or **VC-DomLE**). This shows the proposed rule induction accelerator save time remarkably.
- Furthermore, it is observed from Figs.6.2-6.3 that the gaps between the trendlines of each pair become profoundly larger as the instances increases, which demonstrates that the proposed accelerator works remarkably efficiently compared to its unaccelerated counterpart.
- It is noted from Fig.6.5 that the gap between the trendlines of **VC-DomLE** and **A-VC-DomLE** is tiny on some large datasets, such as Nursery and Thyroid. This shows that **A-VC-DomLE** is poorly accelerated as it spends most time on backward-deleting rule extraction, and less time on rule induction. Thus, our accelerated rule induction could not expedite the computation time of **A-VC-DomLE** substantially.
- Tables 6.2-6.5 indicate that the accelerated algorithm is much faster than their unaccelerated counterparts. Furthermore, it is observed that the induced rule set obtained by each accelerated algorithm is the same as that produced by the unaccelerated one, which benefits from rank preservation property. Hence, rule classifiers based on the accelerated rule induction and the uncelebrated one have the same solutions.
- Finally, Tables 6.2-6.5 show that among four types of rough-based classifiers, the proposed BSRC is the fastest.

TABLE 6.2
TIME AND INDUCED RULES OF THE ALGORITHMS CVRC AND A-CVRC

| Dataset | No. Objects | CVRC | | A-CVRC | | Reduced time(s) |
| --- | --- | --- | --- | --- | --- | --- |
| | | Time(s) | No. Rules | Time(s) | No. Rules | |
| QSAR | 1,055 | 542 | 180 | 177 | 180 | 365 |
| Segm | 2,310 | 451 | 146 | 155 | 146 | 296 |
| Spam | 4,601 | 5,505 | 384 | 1,218 | 384 | 4,286 |
| texture | 5,500 | 6,698 | 179 | 1,430 | 179 | 5,268 |
| Park | 5,875 | 1,775 | 38 | 567 | 38 | 1,208 |
| sat | 6,435 | 11,094 | 514 | 2,242 | 514 | 8,852 |
| nursery | 13,576 | 3,640 | 499 | 1,084 | 499 | 2,556 |
| thyroid | 7,200 | 1,719 | 211 | 535 | 211 | 1,184 |

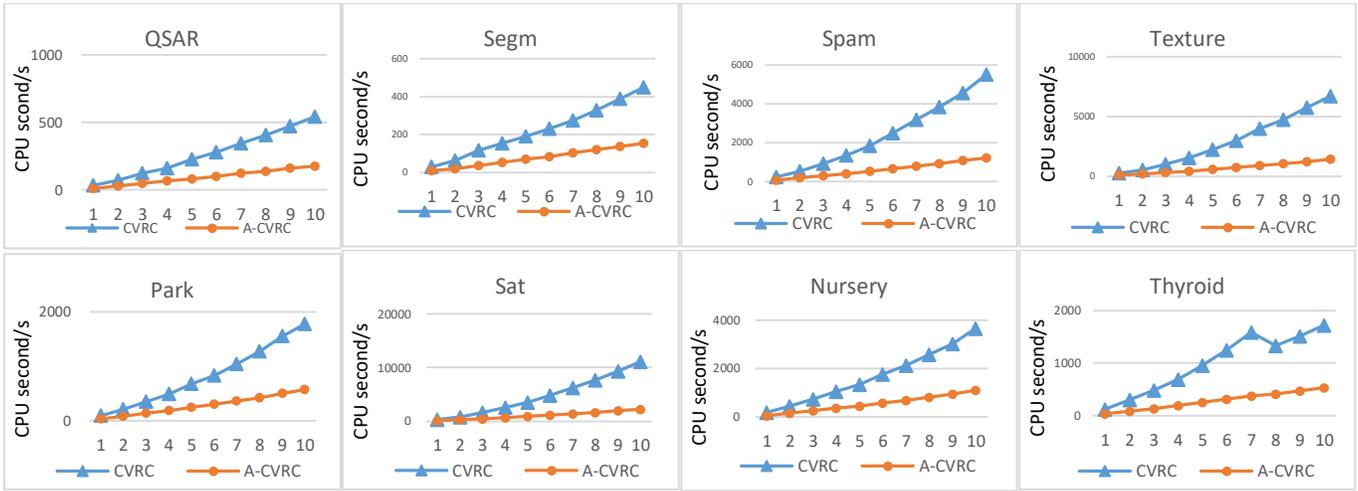

Fig. 6.2: Execution times of CVRC and A-CVRC versus the size of data. In each sub-figure, the x,y-coordinate pertain to dataset size, and the running time(/s).

TABLE 6.3
TIME AND INDUCED RULES OF THE ALGORITHMS BSRC AND A-BSRC

| Dataset | No. Objects | BSRC Time(s) | BSRC No. Rule | A-BSRC Time(s) | A-BSRC No. Rule | Reduced time |
|---|---|---|---|---|---|---|
| QSAR | 1,055 | 126 | 222 | 47 | 223 | 79 |
| Segm | 2,310 | 67 | 183 | 29 | 179 | 38 |
| Spam | 4,601 | 977 | 553 | 249 | 553 | 727 |
| texture | 5,500 | 922 | 212 | 251 | 212 | 671 |
| Park | 5,875 | 360 | 79 | 142 | 79 | 218 |
| sat | 6,435 | 1,641 | 684 | 410 | 685 | 1,231 |
| nursery | 13,576 | 466 | 528 | 265 | 528 | 201 |
| thyroid | 7,200 | 345 | 618 | 79 | 618 | 266 |

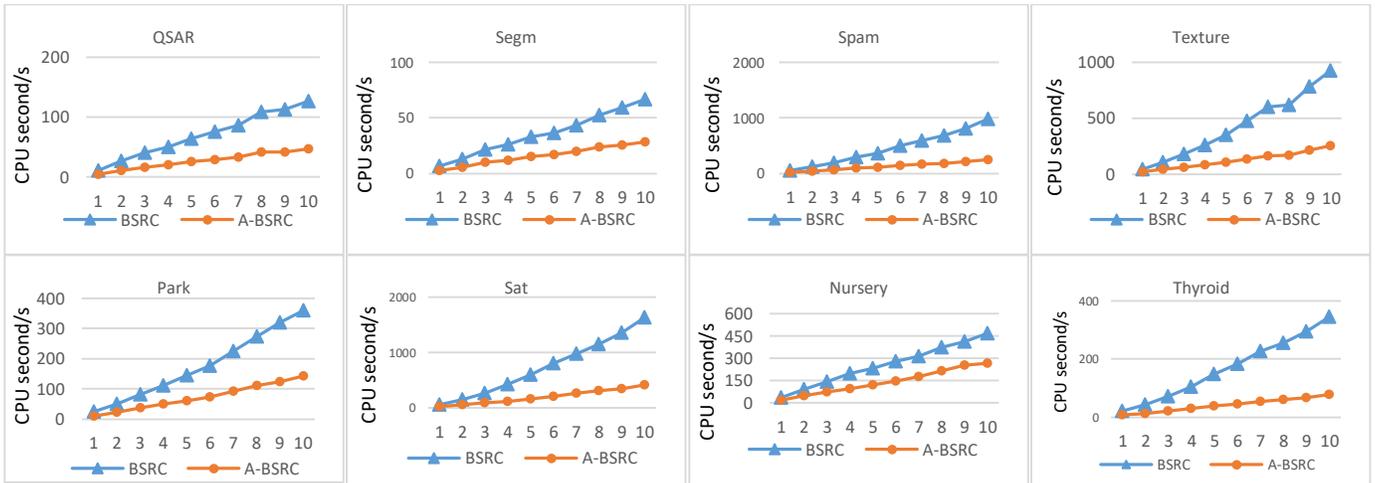

Fig. 6.3: Execution times of BSRC and A-BSRC versus the size of data. In each sub-figure, the x,y-coordinate pertain to dataset size, and the running time(/s).

TABLE 6.4
TIME AND INDUCED RULES OF THE ALGORITHMS LEM2 AND A-LEM2

| Dataset | No. Objects | LEM2 Time(s) | LEM2 No. Rule | A-LEM2 Time(s) | A-LEM2 No. Rule | Reduced time |
|---|---|---|---|---|---|---|
| QSAR | 1,055 | 517 | 119 | 241 | 119 | 276 |
| Segm | 2,310 | 363 | 104 | 171 | 104 | 193 |
| Spam | 4,601 | 4,974 | 235 | 1,819 | 235 | 3,155 |
| texture | 5,500 | 5,001 | 76 | 1,346 | 76 | 3,655 |
| Park | 5,875 | 1,650 | 14 | 791 | 14 | 858 |
| Sat | 6,435 | 8,408 | 266 | 2,302 | 266 | 6,106 |
| nursery | 13,576 | 2,883 | 275 | 2,205 | 275 | 677 |
| thyroid | 7,200 | 2,682 | 184 | 1,694 | 184 | 988 |

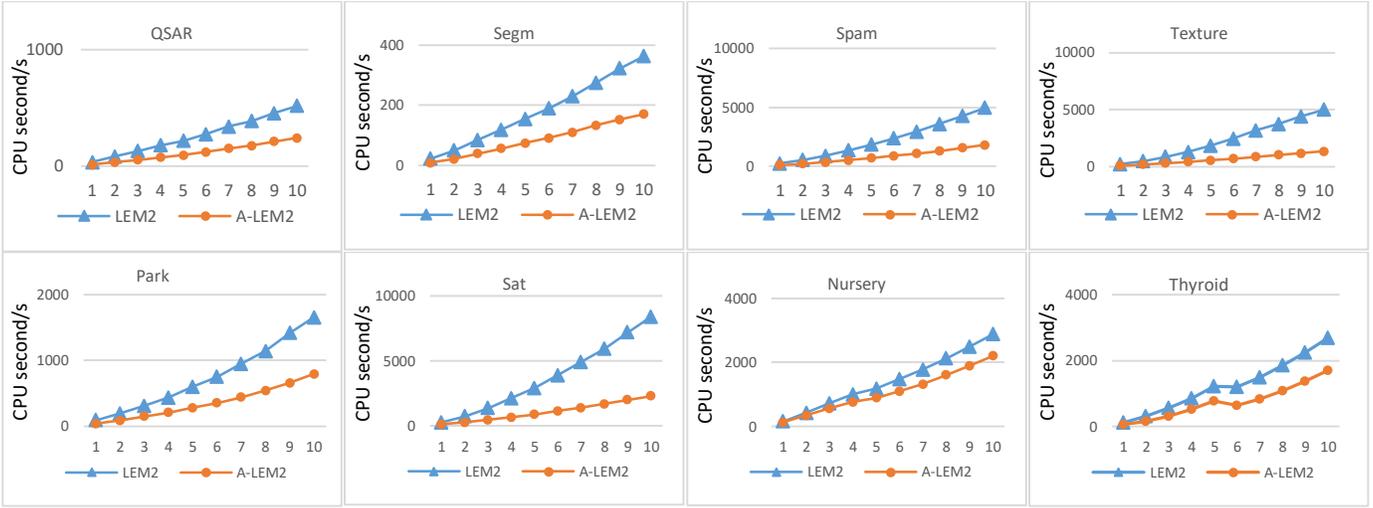

Fig. 6.4: Execution times of LEM2 and A-LEM2 versus the size of data. In each sub-figure, the x,y-coordinate pertain to dataset size, and the running time(/s).

TABLE 6.5
TIME AND INDUCED RULES OF THE ALGORITHMS **VC-DOMLE** AND **A-VC-DOMLE**

| Dataset | No. Objects | VC-DomLE | | A-VC-DomLE | | Reduced time |
| --- | --- | --- | --- | --- | --- | --- |
|  |  | Time(s) | No. Rule | Time(s) | No. Rule |  |
| QSAR | 1,055 | 876 | 175 | 603 | 175 | 273 |
| Cont | 2,310 | 1,025 | 580 | 983 | 580 | 42 |
| Segm | 4,601 | 517 | 151 | 323 | 151 | 194 |
| Spam | 5,500 | 9,424 | 381 | 6,223 | 381 | 3,200 |
| texture | 5,875 | 5,507 | 199 | 1,823 | 199 | 3,684 |
| Park | 6,435 | 3,884 | 89 | 3,015 | 89 | 869 |
| Sat | 13,576 | 11,849 | 551 | 5,753 | 551 | 6,096 |
| nursery | 7,200 | 9,276 | 498 | 8,775 | 498 | 501 |
| thyroid | 1,055 | 12,226 | 193 | 11,283 | 193 | 943 |

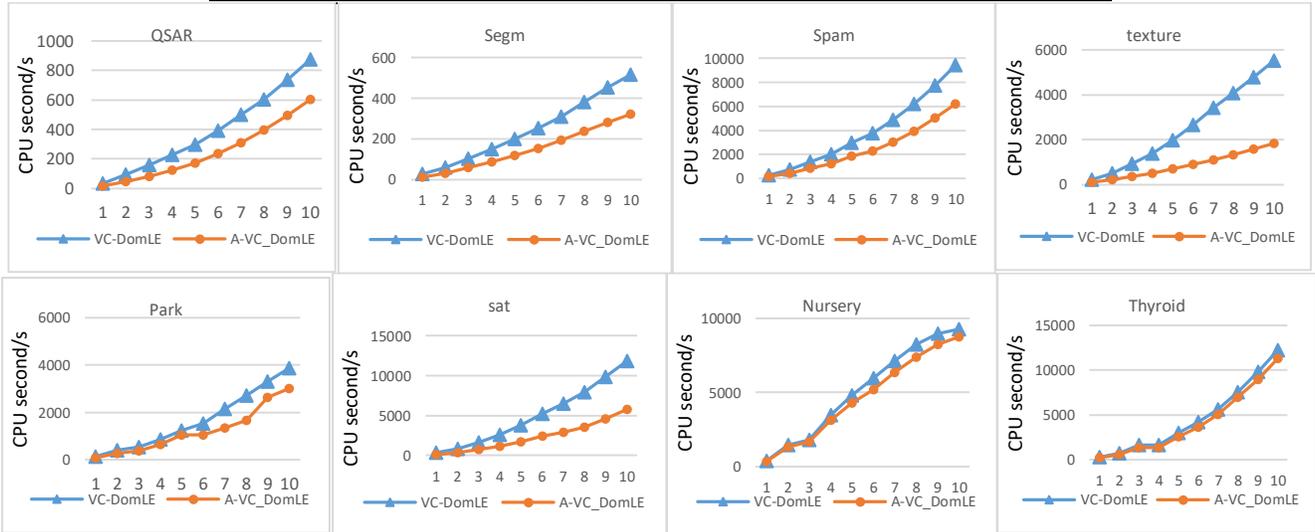

Fig. 6.5: Execution times of **VC-DomLE** and **A-VC-DomLE** versus the size of data. In each sub-figure, the x,y-coordinate pertain to dataset size, and the running time(/s).

### D. Classification Performance Evaluations

This subsection evaluated the classification performance of the proposed algorithms. We compared the proposed accelerated rule classifier with the state-of-the-art rough-based rule classifier (presented in Table 6.6). Furthermore, we compared the proposed rule classifier with some known explainable classifiers (presented in Table 6.7). In Tables 6.6&6.7, the best value of each dataset is in bold font. '●' and '○' respectively denote that A-BSRC was significantly better or worse than the given rule classifier using a t-test with a confidence level of 0.05. The 'A-BSRC: win/tie/loss' row lists the numbers of datasets for which A-BSRC was significantly better or tie or worse than the corresponding method using the t-test.

*1) Compare with Rough-Based Rule Classifiers*

In this part, **A-CVRC** and **A-BSRC were** compared with the representative rough-based rule classifiers, **GFRC**, **A-LEM2** and **A-VC-DomLE** (presented in Table 6.6). We observe the following facts.

- **A-CVRC** and **A-BSRC** outperform other rough-based classifiers on 11+7 out of 20 datasets. This shows that the cooperation of consistence-degree-based rule induction and forward-adding rule extraction can obtain the classifier performing best. Thus, it is necessary to accelerate such classifiers with excellent performance so that they can work efficiently in real scenarios.

- **A-BSRC** outperforms the other rough-based rule classifiers, such as **GFRC**, **A-LEM2** and **A-VC-DomLE** in most cases. This indicates that our proposed method **A-BSRC** is superior among the rough-based rule classifiers. In addition, **A-BSRC** is the only method that is workable on all the twenty datasets. This shows that our proposed method is practical in real applications. Hereinafter we mainly adopt **A-BSRC** to compare with other algorithms.

TABLE 6.6
Mean and standard deviation results for the accuracy of five rule classifiers.

| Datasets | A-BSRC | | A-CVRC | | GFRC | | A-VC-DomLE | | A-LEM2 | |
|---|---|---|---|---|---|---|---|---|---|---|
| | mean | std | mean | std | mean | std | mean | std | mean | std |
| Iono | 0.906 | 0.017 | **0.914** | 0.022 | 0.886 | 0.020 | 0.895 | 0.035 | 0.701 ● | 0.037 |
| Libras | **0.921** | 0.022 | 0.888 | 0.057 | 0.886 | 0.047 | 0.880 | 0.062 | 0.738 ● | 0.071 |
| QSAR | **0.812** | 0.018 | 0.802 | 0.016 | 0.747 ● | 0.024 | 0.808 | 0.022 | 0.648 ● | 0.048 |
| Cont | 0.518 | 0.007 | 0.519 | 0.007 | 0.404 ● | 0.022 | **0.527** | 0.014 | 0.508 | 0.020 |
| Segm | 0.928 | 0.004 | **0.933** | 0.011 | 0.929 | 0.017 | 0.926 | 0.017 | 0.650 ● | 0.045 |
| Spam | 0.909 | 0.007 | 0.909 | 0.013 | 0.846 ● | 0.007 | **0.913** | 0.007 | 0.736 ● | 0.034 |
| texture | **0.953** | 0.005 | 0.951 | 0.003 | 0.920 ● | 0.004 | 0.950 | 0.004 | 0.722 ● | 0.029 |
| optdigits | **0.953** | 0.004 | 0.943 ● | 0.006 | 0.932 ● | 0.007 | 0.942 ● | 0.003 | 0.566 ● | 0.023 |
| Park | 0.991 | 0.002 | **0.996** ○ | 0.002 | 0.986 ● | 0.003 | 0.990 | 0.002 | 0.682 ● | 0.015 |
| sat | **0.866** | 0.012 | 0.858 | 0.009 | 0.846 ● | 0.009 | 0.865 | 0.006 | 0.539 ● | 0.013 |
| Musk2 | **0.968** | 0.003 | **0.968** | 0.006 | 0.921● | 0.003 | 0.967 | 0.005 | 0.813 ● | 0.028 |
| thyroid | 0.933 | 0.004 | **0.940** | 0.005 | 0.906 ● | 0.008 | **0.940** | 0.006 | 0.913 ● | 0.004 |
| ring | 0.942 | 0.003 | 0.940 | 0.004 | 0.915 ● | 0.007 | **0.954** ○ | 0.005 | 0.751 ● | 0.037 |
| coil2000 | 0.931 | 0.004 | 0.926 | 0.006 | 0.921 ● | 0.005 | **0.933** | 0.005 | 0.891 ● | 0.005 |
| crowd | **0.893** | 0.002 | 0.891 | 0.007 | 0.848● | 0.008 | 0.883 | 0.009 | 0.515 ● | 0.025 |
| pendigits | 0.971 | 0.003 | **0.972** | 0.004 | 0.962● | 0.005 | 0.968 | 0.005 | 0.678 ● | 0.015 |
| nursery | 0.952 | 0.002 | 0.950 | 0.003 | **0.959** | 0.004 | 0.951 | 0.005 | 0.801 ● | 0.027 |
| eeg | **0.770** | 0.007 | 0.764 | 0.005 | 0.698● | 0.010 | >24hs | -- | 0.603 ● | 0.019 |
| Shuttle | **0.999** | 0.000 | **0.999** ● | 0.000 | Out of Memory | -- | >24hs | -- | >24hs | -- |
| Sensorless | **0.952** | 0.002 | >24hs | -- | Out of Memory | -- | >24hs | -- | >24hs | -- |
| Count of the best | 11 | | 7 | | 1 | | 5 | | 0 | |
| Count of working | 20 | | 19 | | 18 | | 17 | | 18 | |
| A-BSRC: win\tie\loss | | | 2\16\1 | | 14\4\0 | | 1\15\1 | | 17\1\0 | |

TABLE 6.7
Mean and standard deviation results for the accuracy of A-BSRC and five explainable classifiers.

| Datasets | A-BSRC | | DT | | LR | | GaussianNB | | LDA | | SVM | |
|---|---|---|---|---|---|---|---|---|---|---|---|---|
| | mean | std | mean | std | mean | std | mean | std | mean | std | mean | std |
| Iono | **0.906** | 0.017 | 0.895 | 0.021 | 0.889 | 0.010 | 0.889 | 0.019 | 0.872● | 0.029 | **0.937** | 0.023 |
| Libras | **0.921** | 0.022 | 0.870 | 0.046 | 0.791● | 0.054 | 0.728● | 0.034 | 0.794● | 0.027 | **0.885** | 0.022 |
| QSAR | 0.812 | 0.018 | 0.815 | 0.017 | 0.838 | 0.016 | 0.700● | 0.023 | **0.860**○ | 0.020 | **0.855**○ | 0.010 |
| Cont | 0.518 | 0.007 | 0.506 | 0.028 | 0.521 | 0.038 | 0.484 | 0.034 | 0.524 | 0.048 | **0.530** | 0.031 |
| Segm | 0.928 | 0.004 | **0.964**○ | 0.008 | 0.905● | 0.015 | 0.796● | 0.007 | 0.914 | 0.012 | **0.936** | 0.012 |
| Spam | 0.909 | 0.007 | 0.910 | 0.013 | 0.889● | 0.007 | 0.816● | 0.011 | 0.887● | 0.007 | **0.934**○ | 0.012 |
| texture | 0.953 | 0.005 | 0.928● | 0.008 | 0.972○ | 0.003 | 0.775● | 0.012 | **0.995**○ | 0.001 | **0.991**○ | 0.004 |
| optdigits | 0.953 | 0.004 | 0.900● | 0.004 | **0.970**○ | 0.002 | 0.774● | 0.018 | 0.953 | 0.002 | **0.988**○ | 0.003 |
| Park | **0.991** | 0.002 | 0.986● | 0.003 | 0.755● | 0.006 | 0.686● | 0.008 | 0.812● | 0.009 | 0.953● | 0.005 |
| sat | 0.866 | 0.012 | 0.855 | 0.006 | 0.843● | 0.004 | 0.796● | 0.011 | 0.839● | 0.007 | **0.899**○ | 0.006 |
| Musk2 | **0.968** | 0.003 | 0.967 | 0.006 | 0.936● | 0.006 | 0.839● | 0.013 | 0.944● | 0.007 | 0.964 | 0.002 |
| thyroid | 0.933 | 0.004 | **0.997**○ | 0.001 | 0.937 | 0.001 | 0.118● | 0.014 | **0.938** | 0.002 | 0.937 | 0.001 |
| ring | 0.942 | 0.003 | 0.877● | 0.008 | 0.759● | 0.008 | **0.980**○ | 0.003 | 0.763● | 0.009 | **0.978**○ | 0.003 |
| coil2000 | 0.931 | 0.004 | 0.911● | 0.006 | **0.952**○ | 0.000 | 0.113● | 0.015 | 0.945○ | 0.002 | **0.952**○ | 0.000 |
| crowd | **0.893** | 0.002 | 0.870● | 0.004 | 0.883● | 0.005 | 0.833● | 0.006 | 0.873● | 0.006 | **0.944**○ | 0.004 |
| pendigits | 0.971 | 0.003 | 0.960● | 0.004 | 0.936● | 0.006 | 0.856● | 0.008 | 0.876● | 0.007 | **0.994**○ | 0.001 |
| nursery | 0.952 | 0.002 | **0.997**○ | 0.001 | 0.514● | 0.004 | 0.655● | 0.006 | 0.520● | 0.005 | 0.922● | 0.007 |
| eeg | **0.770** | 0.007 | **0.834**○ | 0.006 | 0.552● | 0.001 | 0.454● | 0.009 | 0.641● | 0.012 | 0.551● | 0.000 |
| Shuttle | 0.999 | 0.000 | **1.000**○ | 0.000 | 0.970● | 0.001 | 0.166● | 0.206 | 0.944● | 0.002 | 0.998● | 0.000 |
| Sensorless | **0.952** | 0.002 | 0.984○ | 0.001 | 0.829● | 0.004 | 0.734● | 0.016 | 0.852● | 0.005 | 0.901● | 0.002 |
| Count of the top 2 | 11 | | 9 | | 2 | | 1 | | 4 | | 13 | |
| A-BSRC: win\tie\loss | | | 7\7\6 | | 13\4\3 | | 17\2\1 | | 13\4\3 | | 5\6\9 | |

### 2) Compare with Explainable Classifiers

In machine learning (ML) systems, interpretability, also called explainability, is defined as the ability to explain or to present in understandable terms to a human [45]. Furthely, [19] stated that an explainable classifier can provide qualitative understanding between the input variables and the response. From the viewpoints of explainablity, the classifiers in ML systems are roughly split into explainable and unexplainable. Just as mentioned in [19,45], some linear models (such as logistic regression, linear discriminant analysis), decision tree, K-nearest neighbor, SVM and rule lists are such explainable classification models. As an explainable classifier, it is unreasonable to compare our proposed rule-based classifier with unexplainable ones, such as deep neural networks. Thus, in this study, different type of explainable classifiers, such as decision tree (DT), linear regression (LR), GaussianNB, linear discriminant analysis (LDA) and support vector machine (SVM), were compared with **A-BSRC**. The comparison results presented in Table 6.7 reveal the following facts.

- **A-BSRC** outperforms LR, GaussianNB, LDA and DT in most cases. This shows that **A-BSRC** works more effectively than most of the existing explainable classifiers.
- As rule-based classifier, **A-BSRC** is not inferior to another rule-based classifier **DT** on 14 out of 20 datasets. This shows that **A-BSRC** is comparable or even outperforms **DT**.

- **A-BSRC** is not inferior to **SVM** on 11 out of 20 datasets. This shows that the performance of **A-BSRC** is comparable to **SVM** in some cases. The superiority of **A-BSRC** over **SVM** is that some human-understanding decision rules are obtained. Thus **A-BSRC** is fit for the applications of decision making and so on.

*E. Discussions*

This study aims at developing an efficient accelerating approach for solving rule induction on large-scale datasets. Large-scale datasets may be with high dimension or large size of instances. To better understand the sensitivity and stability of model results to the changes in the values of dimensionality and instance size, the analysis is performed on two basic parameters, namely, the number of features and the number of instances in classification system.

Efficiency for building classifier in a large-scale classification system is the critical factor. In this regard, the execution time of accelerated and unaccelerated algorithms on dataset 'Musk' was illustrated in Fig.6.6; they can offer some practical implications and useful managerial insights.

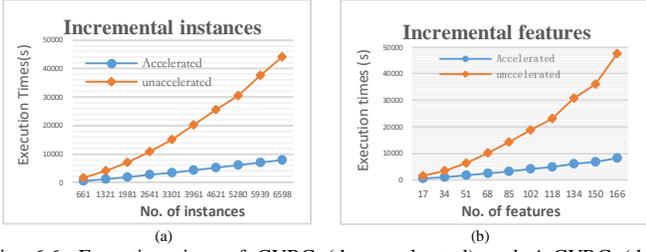

Fig. 6.6: Execution time of CVRC (the accelerated) and A-CVRC (the unaccelerated) versus the size of features (a) and the size of instances on 'Musk'.

Fig.6.6 (a) shows that the execution time of accelerated algorithm is a slightly increasing function of the number of features, whereas the unaccelerated one increases remarkably. For example, the running time of the accelerated on the set with 6598 instances saves more than 30,000s compared to the unaccelerated; whereas the running time on the set with 661 instances saves far less than 2,000s. This result indicates that the more the instances are, the more execution time our accelerator can save. Also, Fig.6.6 (b) shows the similar trend with the incremental features, which indicates that the more the features are, the more execution time our accelerator can save. Thus, it is supposed that our accelerator is efficient and practical to not only the data with large-size instances, but also those with high dimension.

## VII. Conclusions

To overcome the scalability limitations of the existing fuzzy rough-based rule induction scheme, this study developed a general accelerated framework based on FRS theory. Accordingly, we first proposed an alternative to rule induction, **CAR**, which is computationally intensive but can form the basis for designing an accelerator for rule induction. Then, Key Set is proposed to reveal those informative instances for updating the rule induction, which can work as the compacted search space. Finally, rule induction is accelerated on this compacted search space. More importantly, the strict mathematical foundation, such as the monotonicity of Key Set and Discernible Set, and rank preservation property, ensures that the induced rules achieved by the accelerator are the same as those achieved by the unaccelerated one. Additionally, extensive experiments revealed that the proposed accelerator vastly decreased the execution time with few or even no classification performance loss.

The most significant advantage of our proposed method is its efficiency in dealing with classifier building on large-size datasets. Owing to the finding of Key Set and Discernible Set, our proposed method can accelerate classifier building based on the FRS theory by effectively compacting the search space. However, the main limitation of the proposed method is that it has been designed for static data and is not suitable for dynamic datasets with streaming instances. Interestingly, it is feasible to incrementally update Key Set and Discernible Set in dynamic decision table. Thus, in future study, we plan to extend our acceleration idea into dynamic circumstances by cooperated with incremental learning.

## Appendix I: Triangular Operators

We present and exemplify some notions of fuzzy logical operators [4], i.e., triangular norm (or $T$–norm), triangular conorm (or $T$–conorm), negator, dual and $T$–residuated implication and its dual operation, which will be used to construct the set approximation operators in the FRS.

A triangular norm, or shortly $T$–norm, is a function $T:[0,1] \times [0,1] \to [0,1]$ that satisfies: Monotonicity (if $< \alpha, y < \beta$, then $T(x,y) \leq T(\alpha,\beta)$ ); Commutativity ( $T(x,y) = T(y,x)$ ); Associativity ( $T(T(x,y),z) = T(x,T(y,z))$ ), Boundary condition ( $T(x,1) = x$ ). The most typical continuous $T$–norms include the standard *min* operator (the largest $T$–norm) $T_M(x,y) = min\{x,y\}$, algebraic product $T_P(x,y) = x \cdot y$ and the bounded intersection (also called the Lukasiewicz $T$–norm) $T_L(x,y) = max\{0, x+y-1\}$.

A triangular conorm, or shortly $T$–conorm, is an increasing, commutative and associative function $S:[0,1] \times [0,1] \to [0,1]$ that satisfies the boundary condition: $\forall x \in [0,1], S(x,0) = x$.

A negator $N$ is a decreasing function $N:[0,1] \to [0,1]$ that satisfies $N(0) = 1$ and $N(1) = 0$.

Given a lower semi-continuous triangular norm $T$, the residuation implication, also known as the $T$–residuated implication, is a function $\vartheta:[0,1] \times [0,1] \to [0,1]$ that satisfies $\vartheta(x,y) = sup\{z|z \in [0,1], T(x,z) \leq y\}$ for every $x,y \in [0,1]$. The Lukasiewicz implication $\vartheta_L$ which is based on $T_L$: $\vartheta_{T_L} = min\{1-x+y,1\}$.

## Appendix II Two Types of Rule Extraction Strategies.

From the viewpoint of search strategies, the rule extraction methods are split into forward-adding and backward-deleting.

The search strategy of [7] comes under the forward-adding type. We present it in Steps 4-8 of Algorithm A.1, designed to extract a near-minimal rule set. And this strategy is called rule extraction in **GFRC**.

| **Algorithm A.1.** To extract a near-minimal rule set by **GFRC** [7] |
|---|
| **Input:** $FD = (U, C \cup D); \alpha \in [0,1)$; |
| **Output**: Near-minimal rule set: $minimal\_rule\_set$; |
| **Step 1: Calculate** the induced rule $reduct(x)$ by **Algorithm 2.1**, **DVR**; |
| **Step 2: Add** all $reduct(x)$ into $all\_rules$, $minimal\_rule\_set \leftarrow \emptyset$; |
| **Step 3: Calculate** the cover degree of every rule in $all\_rules$; |
| **Step 4: While** $all\_rules$ is not empty, **do** |
| **Step 5:**  Add the rule $Rule(x^*)$ which has the maximum cover degree into $minimal\_rule\_set$; |

><  14

**Step 6:** Delete rules cover by $Rule(x^*)$ from $all\_rules$;
**Step 7:** Update the cover degree of every rule in $all\_rules$ by deleting those instances covered by $Rule(x^*)$;
**Step 8: End while**
**Step 9: Return** $minimal\_rule\_set$.

Here, "$all\_rules$" denotes the collection of the induced rules; "$Rule(x)$" denotes the induced rule of $x$; "$cover\_degree(x)$" denotes the number of rules covered by the "$Rule(x)$.

A basic deleting strategy was proposed in LEM2[12], seen as Algorithm A.2, in which the key idea is, if one rule is covered by a certain rule in the candidate of the minimal rule-set, the former rule can be deleted.

---

**Algorithm A.2.** Rule extraction in **LEM2** [12]

**Input:** 1) $all\_rules$: the collection of the induced rules; 2) $Rule(x)$: one induced rule of $x$;
**Output:** $minimal\_rule\_set$: the extracted the near-minimal rule set;
**Step 1: For every rule** $rule(x)$ in $all\_rules$, **do**
**Step 2:** $rest\_rule \leftarrow all\_rules - rule(x)$;
**Step 3:** For every $rule(y)$ in $rest\_rule$, **do**
**Step 4:** If the rule $rule(x)$ is covered by the rule $rule(y)$,
**Step 5:** $all\_rules \leftarrow all\_rules - rule(x)$;
**Step 6:** Break;
**Step 7:** End for
**Step 8: End for**
**Step 9:** $minimal\_rule\_set \leftarrow all\_rules$;
**Step 10: Return** $minimal\_rule\_set$.

---

Another backward-deleting search strategy is the improved version of **LEM2**. The key idea is that if the covering instance set of one rule is covered by the union of other covering sets of other rules, then this rule can be deleted [13].

---

**Algorithm A.3.** Rule extraction in **VC-DomLE** [13]

**Input:** 1) $all\_rules$: the collection of the induced rules;
   2) $Rule(x)$: one induced rule of $x$;
**Output:** $minimal\_rule\_set$: the extracted the near-minimal rule set;
**Step 1:** $\Gamma \leftarrow all\ induced\ rules$;
**Step 2: For every** $T \in \Gamma$, **do** (here T is an induced rule)
**Step 3:** If $[T] \subseteq \cup_{S \in \Gamma - T} [S]$,
**Step 4:** let $\Gamma \leftarrow \Gamma - T$; here $[T]$ is the objects covered by $T$;
**Step 5:** End if
**Step 6: End for**
**Step 7:** $minimal\_rule\_set \leftarrow \Gamma$;
**Step 8: Return** $minimal\_rule\_set$.

---

<sup>>15</sup>

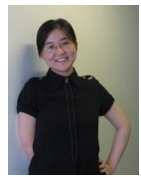
**Suyun Zhao** received the Bachelor's and Master's degrees from Hebei University, Baoding, China, in 2001 and 2004, respectively, and the Ph.D. degree from The Hong Kong Polytechnic University, Hong Kong. She is currently with Renmin University of China, Beijing, China. Her research interests include machine learning, pattern recognition, and uncertain information processing.

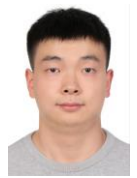
**Zhigang Dai** is currently a master student at Renmin University of China, Beijing, China. He received Bachelor degree from Anhui University, Hefei, China. His research interests include machine learning, Semi-supervised learning and incremental classification.

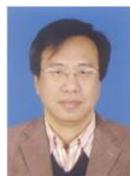
**Xizhao Wang** (M'03–SM'04–F'12) received the Ph.D. degree in computer science from the Harbin Institute of Technology, Harbin, China, in 1998. He is a Professor in Shenzhen University, Shen-zhen, China. His major research interests include uncertainty modeling and machine learning for big data. Dr. Wang is the Chair of the IEEE Systems, Man, and Cybernetics (SMC) Technical Committee on Computational Intelligence, an Editor-in-Chief of the Machine Learning and Cybernetics Journal.

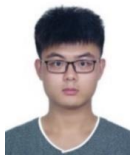
**Peng Ni** is currently a master student at Renmin University of China, Beijing, China. His research interests include machine learning, uncertain artificial intelligence, incremental feature selection.

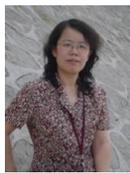
**Hong Chen** received the Bachelor and Master degrees from the Renmin University of China, Beijing, China, in 1986 and 1989, respectively, and the Doctor degree from the Chinese Academy of Sciences, Beijing, in 2000. She is currently with Renmin University of China. Her research interests include high-performance database system, data warehouse and data mining, and data management in wireless sensor networks.

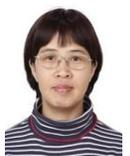
**Cuiping Li** received the B.E. and M.E. degrees from Xi'an Jiao Tong University, Xi'an, China, in 1994 and 1997, respectively, and the Ph.D. degree from the Institute of Computing Technology, Chinese Academy of Sciences, Beijing, China, in 2003. She is a Professor with the Renmin University of China, Beijing. Her current research interests include database systems, data warehousing, and data mining.